%% file: main.tex
\documentclass[10pt,twocolumn,letterpaper]{article}

\usepackage[pagenumbers]{iccv} 

\definecolor{iccvblue}{rgb}{0.21,0.49,0.74}
\usepackage[pagebackref,breaklinks,colorlinks,allcolors=iccvblue]{hyperref}
\usepackage{colortbl}
\input{custom}

\input{math}

\def\paperID{1765} 
\def\confName{ICCV}
\def\confYear{2025}

\title{Toward Material-Agnostic System Identification from Videos}

\author{
\centering
\begin{tabular}{c}
Yizhou Zhao\textsuperscript{1} \quad
Haoyu Chen\textsuperscript{1} \quad
Chunjiang Liu\textsuperscript{1} \quad
Zhenyang Li\textsuperscript{2} \quad
Charles Herrmann\textsuperscript{3} \quad \\
Junhwa Hur\textsuperscript{3} \quad
Yinxiao Li\textsuperscript{3} \quad
Ming-Hsuan Yang\textsuperscript{3,4} \quad
Bhiksha Raj\textsuperscript{1} \quad
Min Xu\textsuperscript{1}\thanks{} \\
\textsuperscript{1}Carnegie Mellon University \quad
\textsuperscript{2}University of Alabama at Birmingham \quad
\textsuperscript{3}Google \quad
\textsuperscript{4}UC Merced \\
\url{https://github.com/Skaldak/MASIV}
\end{tabular}
}

\begin{document}
\twocolumn[{%
\renewcommand\twocolumn[1][]{#1}%
\maketitle
\input{sections/0_teaser}
}]
\renewcommand{\thefootnote}{\fnsymbol{footnote}}
\footnotetext[1]{Corresponding author}
\input{sections/0_abstract}    
\input{sections/1_intro}
\input{sections/2_related}
\input{sections/3_method}

\input{sections/4_experiments}
\input{sections/5_conclusion}
\clearpage
\input{sections/6_acknowledgment}
{
    \small
    \bibliographystyle{ieeenat_fullname}
    \bibliography{main}
}
\input{sections/X_suppl}

\end{document}

%% file: custom.tex
\usepackage{adjustbox}
\usepackage{bm}
\usepackage{multirow}
\usepackage{multicol}
\usepackage{color, colortbl}
\usepackage{pifont}
\usepackage{placeins}
\usepackage{transparent}
\usepackage{algorithm}
\usepackage{algorithmic}
\usepackage{balance}
\usepackage[accsupp]{axessibility}  

\usepackage[capitalize]{cleveref}
\Crefname{section}{Section}{Sections}
\crefname{section}{Sec.}{Secs.}
\Crefname{align}{Equation}{Equations}
\crefname{align}{Eq.}{Eqs.}
\Crefname{equation}{Equation}{Equations}
\crefname{equation}{Eq.}{Eqs.}
\Crefname{figure}{Figure}{Figures}
\crefname{figure}{Fig.}{Figs.}
\Crefname{table}{Table}{Tables}
\crefname{table}{Tab.}{Tabs.}

\newcommand\minisection[1]{\vspace{1mm}\noindent \textbf{#1}}
\newcommand{\norm}[1]{\left\lVert#1\right\rVert}

\newcommand{\cmark}{\ding{51}}
\newcommand{\xmark}{\ding{55}}
\newcommand{\model}{\mbox{MASIV}\xspace}

\newcommand{\supp}{\mbox{Supp.~Mat.}\xspace}

\definecolor{Gray}{gray}{0.9}
\definecolor{Celadon}{rgb}{0.67, 0.88, 0.69}
\definecolor{Cream}{rgb}{1.0, 0.99, 0.82}

\definecolor{gold}{HTML}{FAE37F}
\definecolor{silver}{HTML}{D7D7D7}
\definecolor{bronze}{HTML}{EDBA91}

\makeatletter
\newcommand*{\@rowstyle}{}
\newcommand*{\rowstyle}[1]{
  \gdef\@rowstyle{#1}%
  \@rowstyle\ignorespaces%
}
\newcolumntype{=}{
  >{\gdef\@rowstyle{}}%
}
\newcolumntype{+}{
  >{\@rowstyle}%
}
\makeatother

%% file: math.tex
\newcommand{\Dc}{\mathcal{D}}
\newcommand{\Ec}{\mathcal{E}}

\newcommand{\Gc}{\mathcal{G}}

\newcommand{\Ic}{\mathcal{I}}

\newcommand{\Lc}{\mathcal{L}}
\newcommand{\Mc}{\mathcal{M}}

\newcommand{\Pc}{\mathcal{P}}

\newcommand{\Sc}{\mathcal{S}}

\newcommand{\Rb}{\mathbb{R}}

\newcommand{\cv}{\mathbf{c}}

\newcommand{\fv}{\mathbf{f}}

\newcommand{\pv}{\mathbf{p}}
\newcommand{\qv}{\mathbf{q}}

\newcommand{\sv}{\mathbf{s}}

\newcommand{\uv}{\mathbf{u}}
\newcommand{\vv}{\mathbf{v}}
\newcommand{\wv}{\mathbf{w}}
\newcommand{\xv}{\mathbf{x}}

\newcommand{\Bv}{\mathbf{B}}
\newcommand{\Cv}{\mathbf{C}}
\newcommand{\Dv}{\mathbf{D}}

\newcommand{\Fv}{\mathbf{F}}

\newcommand{\Iv}{\mathbf{I}}

\newcommand{\Mv}{\mathbf{M}}

\newcommand{\Pv}{\mathbf{P}}

\newcommand{\Rv}{\mathbf{R}}
\newcommand{\Sv}{\mathbf{S}}

\newcommand{\muv        }{\boldsymbol \mu        }

\newcommand{\Sigmav     }{\boldsymbol \Sigma     }

%% file: sections/0_teaser.tex
\begin{center}
    \centering
    \vspace{-1.5em}
    \includegraphics[width=.9\linewidth]{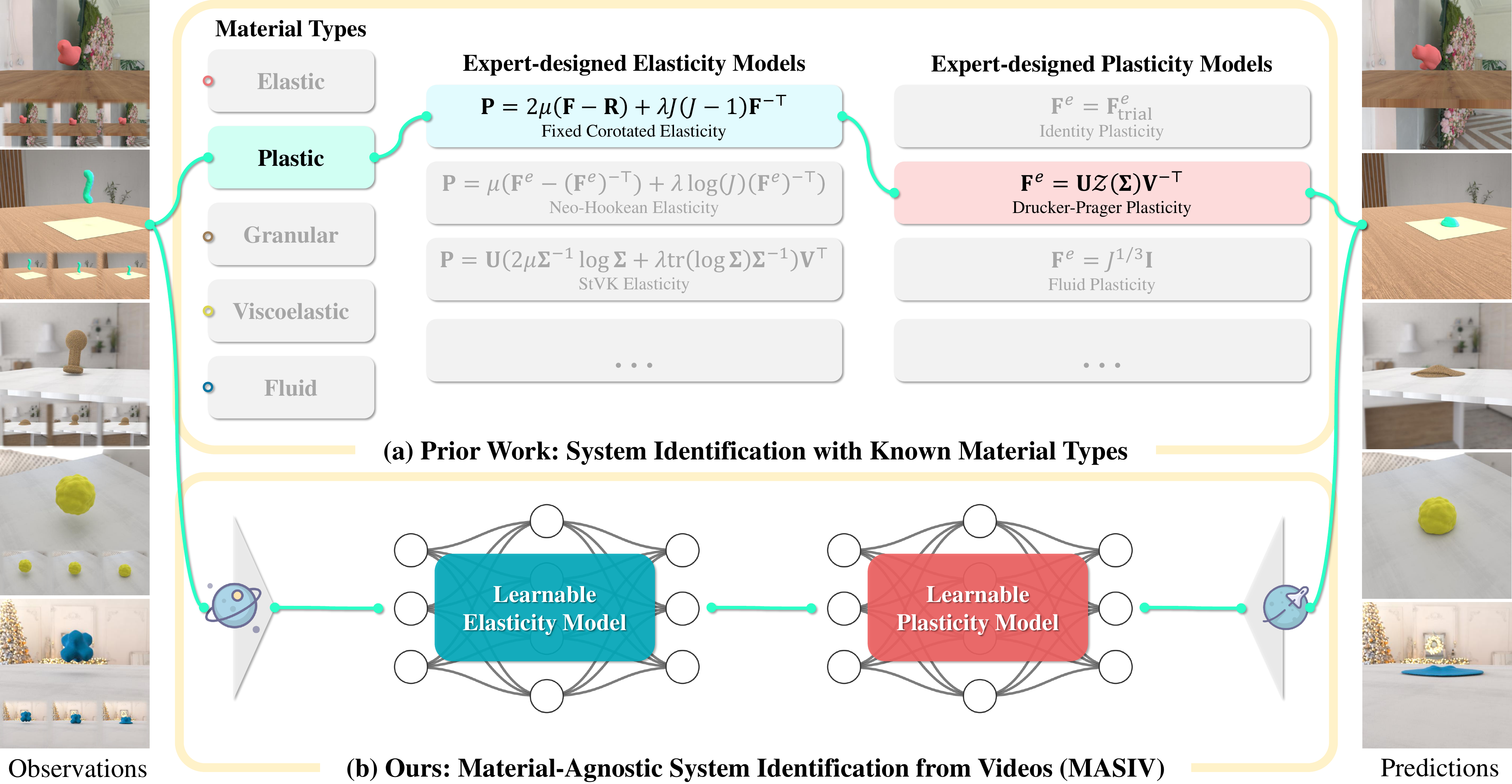}
    \captionsetup{type=figure}
    \caption{System identification from video observations seeks to understand the intrinsic dynamics of a given scene. (a) Existing methods rely on scene-specific material priors. (b) Our \model removes this requirement, grounding object motion in a material-agnostic manner.}
    \label{fig:teaser}
\end{center}

%% file: sections/0_abstract.tex
\begin{abstract}
System identification from videos aims to recover object geometry and governing physical laws. 
Existing methods integrate differentiable rendering with simulation but rely on predefined material priors, limiting their ability to handle unknown ones.
We introduce \model, the first vision-based framework for material-agnostic system identification. Unlike existing approaches that depend on hand-crafted constitutive laws, \model employs learnable neural constitutive models, inferring object dynamics without assuming a scene-specific material prior.
However, the absence of full particle state information imposes unique challenges, leading to unstable optimization and physically implausible behaviors.
To address this, we introduce dense geometric guidance by reconstructing continuum particle trajectories, providing temporally rich motion constraints beyond sparse visual cues.
Comprehensive experiments show that \model achieves state-of-the-art performance in geometric accuracy, rendering quality, and generalization ability.
\end{abstract}

%% file: sections/1_intro.tex
\section{Introduction}
\label{sec:intro}
From the flow of water down a stream to the deformation of a rubber ball upon impact, the physical world operates with a set of underlying principles that govern motion and interaction.
Humans intuitively grasp these principles through observation, allowing us to see a scene and instinctively predict its outcome or re-imagine it with different initial dynamics. Material identification forms the basis of this ability, requiring us to determine an object's composition and understand how that influences its response to forces. This capability is fundamental to world modeling, the process of constructing internal representations that help us understand and anticipate the evolution of the environment. However, replicating the ability to generalize physical laws to diverse scenarios remains a challenge for artificial intelligence.

To bridge this gap, researchers have explored vision-based system identification~\cite{chari2019visual, guan2022neurofluid, jatavallabhula2021gradsim, hofherr2023neural, li2023pac, cai2024gic, cao2025neuma}, which aims to infer governing physical laws from visual observations.
Typically, they optimize a parameterized physical model, by end-to-end integrating differentiable renderers, such as Neural Radiance Fields (NeRF)~\cite{mildenhall2021nerf} and 3D Gaussian Splatting (3DGS)~\cite{kerbl20233d}, with differentiable simulators~\cite{jiang2016material, hu2018moving, hu2019difftaichi, sulsky1995application, xue2023jax}.
To enforce physically meaningful constraints on the simulation, existing approaches~\cite{zhong2025reconstruction, li2023pac, cai2024gic} often rely on material-specific constitutive laws, such as elasticity, plasticity, or viscosity models.
These laws define how an object's material properties influence its response to external forces, allowing system identification methods to estimate parameters such as stiffness, damping, or friction.
While effective in controlled settings, this dependence on hand-crafted material models limits adaptability.
Moreover, these methods assume knowing the material type for a given scene, making it difficult for them to generalize across in-the-wild scenarios where material properties are unknown.

Directly targeting these limitations, we introduce Material-Agnostic System Identification from Videos (\model), a vision-based approach that infers object dynamics free of predefined material priors.
As illustrated in~\cref{fig:teaser}, rather than relying on expert-designed elasticity and plasticity models, \model employs fully learnable neural constitutive models inspired by NCLaw~\cite{ma2023learning}.
However, directly applying neural constitutive models to vision-based system identification poses significant challenges.
Similar to the learning process of neural PDE solvers~\cite{brandstetter2022message, herde2025poseidon,li2024physics,lippe2023pde}, the original formulation of NCLaw assumes access to complete state observations of simulated particles, including position, velocity, deformation gradient, and affine momentum.
Since visual observation fails to provide such complete information, simply relying on frame-wise pixel supervision provides insufficient constraints.
This leads to unstable optimization and poor convergence~\cite{lin2025omniphysgs}.

We combine two main insights to enable this optimization.
First, despite the lack of ground truth full particle state, frame-wise observations still allow reasonable estimates for some of them, such as position and velocity.
Existing dynamic reconstruction methods~\cite{luiten2023dynamic,li2024spacetime,lu20243d} factorize moving objects with their canonical representation and corresponding deformations, which enables per-frame position tracking.
Further, the commonly adopted motion basis~\cite{zhou2012moving,mo2023continuous,ye2021motion} provides a low-rank encoding of object motion, facilitating state interpolation between frames.
This formulation allows us to reconstruct particle trajectories over dense time steps.
Second, although trajectory estimations are inherently limited to visible object regions, internal deformations can be inferred by propagating surface motions inward.
Our intuition is that nearby particles inside an object should be deformed similarly, akin to the As-Rigid-As-Possible (ARAP) prior.

We use these insights to guide the design of our framework.
Concretely, we reconstruct dynamic Gaussians from multi-view video inputs, simultaneously establishing geometry and deformation models of the scene.
The geometry model is then transformed into a solid continuum, where the deformation model is fine-tuned to encode both the exterior and the interior motion.
To identify intrinsic dynamics, we guide the learning of the neural constitutive model using two complementary cues: appearance cues from video frames and geometric constraints from the fine-tuned deformation model.
In this way, \model distills dense particle motion from sparse visual observations and guides the learning of neural constitutive models for stable, material-agnostic system identification.

Our contributions can be summed up as follows:
\begin{itemize}
    \item We present a novel framework, \model, that determines the intrinsic dynamics of an object from videos in a material-agnostic manner. In contrast, all other current vision-based system identification methods~\cite{li2023pac,cai2024gic,cao2025neuma} necessitate predefined material priors.
    \item We introduce temporally dense geometry guidance to constrain object behaviors with neural constitutive models, facilitating smoother convergence and improved performance for material-agnostic system identification.
    \item Through extensive qualitative and quantitative experiments, we demonstrate that \model achieves superior performance in both observable state recovery and future state prediction, even without explicit material priors.
\end{itemize}

%% file: sections/2_related.tex
\section{Related Work}
\label{sec:related}
\minisection{Dynamic reconstruction}
is a fundamental task in computer vision, aimed at reconstructing high-fidelity representations of complex dynamic scenes from diverse inputs such as monocular and multi-view videos~\cite{liu2024modgs,girish2024queen,yang2023real,stearns2024dynamic,attal2023hyperreel,cai2024structure,li2023dynibar,wang2023flow,xian2021space,yu2024cogs,yu2023dylin}. Earlier research primarily focused on augmenting canonical Neural Radiance Fields (NeRF)~\cite{mildenhall2021nerf} with a deformation field~\cite{pumarola2021d}, which were later enhanced by incorporating volume-preserving regularization techniques~\cite{park2021nerfies,park2021hypernerf}. While primarily focused on the novel view synthesis task, their neural implicit representations often introduce noisy deformations, limiting the accuracy of recovered geometries required for reliable physics-based property estimation~\cite{li2023pac}. With the explicit representation of the scene using Gaussian ellipsoids, 3D Gaussian Splatting (3DGS)~\cite{kerbl20233d} achieves efficient dynamic scene reconstruction. Building on the principles of 3DGS, several studies have extended its application to 4D dynamic reconstruction by handling each frame separately~\cite{luiten2023dynamic} or modeling the entire scene with a canonical 3D Gaussian point cloud coupled with a robust deformation framework that accurately maps it to the target scene~\cite{yang2024deformable,kratimenos2025dynmf,wu20244d}.

\minisection{Dynamic simulation}
integrates physics principles into 3D representations to generate realistic motion and interactions. A typical approach to synthesizing dynamic 3D scenes involves integrating a 3D generation pipeline with a video generation model~\cite{bahmani20244d, ling2024align, ren2023dreamgaussian4d, singer2023text}. Recent research incorporates 3D Gaussian kernels as both visual and physical representations, embedding Newtonian dynamics into the Gaussian framework to enable seamless rendering and simulation while introducing physical constraints to ensure plausible dynamics~\cite{xie2024physgaussian, liu2024physics3d, lin2025omniphysgs, li2023pac, qiu2024language, borycki2024gasp, zhong2025reconstruction, fu2024sync4d}. Building on this, Gaussian Splashing~\cite{feng2024gaussian} introduces a position-based dynamics framework within 3D Gaussian Splatting, enabling interactions among solids, fluids, and deformable objects. Additionally, motion-conditioned simulation methods~\cite{li2024generative, geng2024motion, wang2024motionctrl, wu2024draganything, shi2024motion} use trajectory-based guidance to predict object motion, showcasing how generative models can enhance dynamic scene synthesis. However, these methods primarily rely on pre-trained generative models, which often lack a fundamental understanding of physics. More recent works leverage neural physics modeling, incorporating video priors and generative models to infer object dynamics, allowing physics-driven motion synthesis without explicit solvers~\cite{zhang2025physdreamer, liu2025physgen, huang2024dreamphysics,feng2024pie,tan2024physmotion}. However, these methods are constrained to specific material types, limiting their generalization across diverse objects. To address this, we introduce a material-agnostic approach that unifies system identification across different object types, ensuring broader applicability beyond predefined materials.

\minisection{System identification}
aims to understand the physical laws governing the 3D world and is an essential task for simulation~\cite{li2024nvfi, liang2019differentiable, raissi2019physics, sundaresan2022diffcloud, li2022diffcloth, zhong2025reconstruction} and robotic manipulation~\cite{shi2023robocook, shi2024robocraft, liang2024real, zheng2024differentiable, qiao2022neuphysics}. However, accurately recovering both object geometries and physical properties remains a significant challenge. Existing methods often assume known geometries and rely on finite element methods (FEM) or mass-spring systems for dynamic simulation~\cite{takahashi2019video, wang2015deformation}, limiting their ability to handle complex, real-world materials. Neural network-based methods~\cite{sanchez2020learning, li2018learning, xu2019densephysnet} offer greater flexibility by leveraging data-driven models for system identification, but they struggle to generalize across diverse material properties and environmental conditions. More recently, differentiable physics-based simulations have emerged as a promising direction~\cite{hu2019difftaichi, huang2021plasticinelab, chen2022virtual, du2021diffpd, geilinger2020add, heiden2021disect, jatavallabhula2021gradsim, ma2022risp, qiao2021differentiable, kaneko2024improving}. However, these methods often require intricate modeling to bridge the gap between simulated and real-world behavior. To improve adaptability, some approaches focus on neural constitutive modeling, embedding learned material properties within physical simulations to refine expert-designed models~\cite{ma2023learning, cao2025neuma}. Other methods leverage the Material Point Method (MPM) and 3D Gaussian-based representations to enhance geometry reconstruction and material property estimation~\cite{li2023pac, cai2024gic, zhong2025reconstruction,shao2024gausim}. Building upon GIC~\cite{cai2024gic}, our method presents a material-agnostic approach that eliminates reliance on predefined material priors. Rather than assuming material properties, our model infers object dynamics directly from visual observations, making it adapt better to a wide range of unknown or heterogeneous materials.

%% file: sections/3_method.tex
\begin{figure*}[t]
  \centering
   \includegraphics[width=\linewidth]{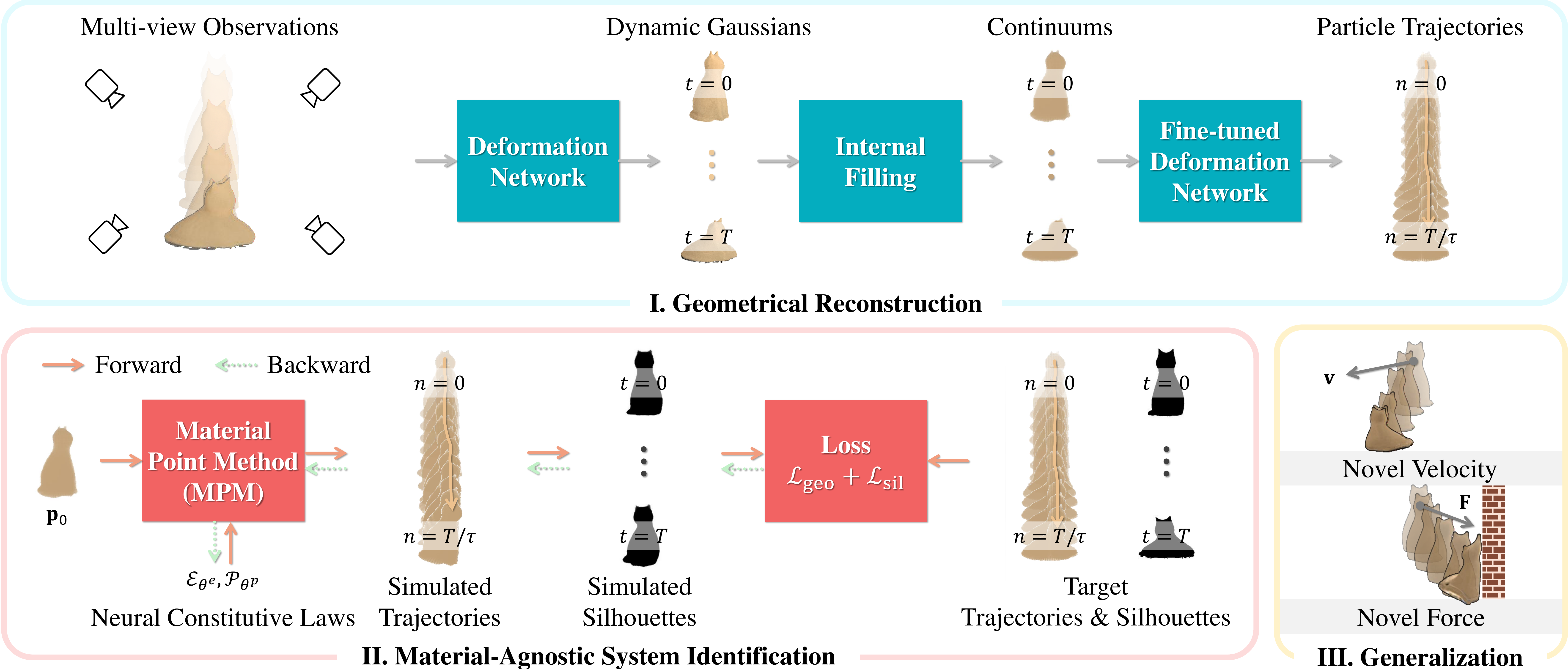}
   \caption{\textbf{The overview of \model.} Our pipeline comprises three phases. Phase I, Geometrical Reconstruction, infers geometrical representations (\cref{subsec:gaussian}) and dense trajectories (\cref{subsec:trajectory}) for the captured object. Phase II, Material-Agnostic System Identification (\cref{subsec:system}), grounds dynamic characteristics by exploiting visual observations with reconstructed motion clues. From this, we obtain a generalizable digital twin in Phase III, which can simulate novel interactions, such as new velocities or forces.}
   \label{fig:pipeline}
\vspace{-1.0em}
\end{figure*}

\section{Method}
Given visual observations $\Iv = \{\Iv_0, \Iv_1, \dots, \Iv_T\}$ over $T$ frames capturing a dynamic object, our goal is to recover its explicit geometric representation and the governing evolution rules of the underlying dynamical system.
Formally, we represent the object's geometry using a set of Gaussian kernels~\cite{kerbl20233d} and integrate neural constitutive laws~\cite{ma2023learning} with the material point method (MPM)~\cite{sulsky1995application, jiang2016material, hu2018moving} to roll out their motion over time.
Unlike~\cite{li2023pac,cao2025neuma,zhong2025reconstruction,cai2024gic}, we do not assume a known material type of the object, while exploring Material-Agnostic System Identification from Videos (\model).
\cref{fig:pipeline} shows an overview of our pipeline.

\subsection{Preliminaries}

\begin{algorithm}[t]
   \caption{MPM state transfer $\Mc$}
   \label{alg:transfer}
\begin{algorithmic}[1]
\INPUT $\sv_n = \{\xv_n,\vv_n,\Fv_n^e\}$
\OUTPUT $\sv_{n+1} = \{\xv_{n+1},\vv_{n+1},\Fv_{n+1}^e\}$
    \STATE for $i=1\dots Q$, $\Pv_n(i)=\Ec(\Fv_n^e(i))$
   \STATE $\xv_{n+1},\vv_{n+1},\Fv_{n+1}^{e,\text{trial}}=\Ic(\xv_n,\vv_n,\Fv_n^e,\Pv_n)$
   \STATE for $i=1\ldots Q$, $\Fv_{n+1}^e(i)=\Pc(\Fv_{n+1}^{e,\text{trial}}(i))$
\end{algorithmic}
\end{algorithm}

\minisection{Material Point Method (MPM)\label{subsubsec:mpm}}
discretizes the object of interest into $Q$ material points and models their evolution over time via a state transfer equation
\begin{align}
\label{eq:transfer}
    \sv_{n+1} = \Mc(\sv_n),\;\forall n=0,1,\dots,N.
\end{align}
$\sv_n$ denotes the particle state at time step $n$. $N$ is the total number of simulation steps with the simulation interval $\tau$ and observed $T$ frames, where $N=T/\tau \gg T$.

The transition process $\Mc$ follows a time-stepping scheme, as outlined in~\cref{alg:transfer}.
The state of particle $i$ at time step $n$ is described by its position $\xv_n(i)$, velocity $\vv_n(i)$, and the elastic component $\Fv^e_n(i)$ of its deformation gradient $\Fv_n(i)$.
At each time step, the elastic constitutive law $\Ec$ computes the first Piola-Kirchhoff stress $\Pv_n(i)$ from the elastic deformation gradient $\Fv_n^e(i)$.  
Subsequently, the time integration scheme $\Ic$ updates all particle states.  
The updated trial elastic deformation gradient, denoted as $\Fv_{n+1}^{e,\text{trial}}(i)$, needs to be further corrected by the plastic constitutive law $\Pc$, which applies plasticity constraints and yields the final elastic deformation gradient $\Fv_{n+1}^e(i)$.
More details of MPM can be found in~\cref{subsec:supp:mpm} of \supp

Existing efforts~\cite{li2023pac,cao2025neuma,zhong2025reconstruction,cai2024gic} for video-based system identification commonly assume known constitutive laws $\Ec$ and $\Pc$, with a particular backbone form, e.g., neo-Hookean elasticity with identity plasticity return function.
This simplifies the problem by estimating a small number of physical properties, such as Young’s modulus, fluid viscosity, friction angles, etc., at the cost of sacrificing flexibility.
In contrast, we explore the same problem without this material-specific assumption.

\subsection{Dynamic Gaussian Reconstruction}
\label{subsec:gaussian}
We start by reconstructing dynamic Gaussians over $T$ frames.
We consider this reconstruction as temporally sparse because the number of simulation steps is hundreds of times more than the number of observed video frames $T$, i.e., $N=T/\tau \gg T$ .
Following~\cite{yang2024deformable, kratimenos2025dynmf, wu20244d, cai2024gic}, we maintain a set of canonical Gaussians $\Gc^*=\{\muv^*, \Sigmav^*, \cv^*, \sigma^*\}$ and use a neural network $\Dc$ to warp them over time to represent observed frames. 
Specifically, a basis network maps a given time step $t$ to $B$ bases, producing deformation basis for position $\Bv^{\muv}(t) \in \Rb^{B \times 3}$ and scale $\Bv^s(t) \in \Rb^{B \times 1}$.  
A coefficient network then estimates the weight of each basis, $\wv(\muv^*,t) \in \Rb^B$, from the center coordinates of canonical Gaussian kernels $\muv^*$ and the time step $t$.  
This results in deformed Gaussian kernels at each time step $t$:
\begin{align}
\label{eq:deform_gaus}
    \muv_t &= \muv^* + \wv(\muv^*,t)\Bv^{\muv}(t), \\
    s_t &= s^* + \wv(\muv^*,t)\Bv^s(t).
\end{align}
We employ the same optimization strategy in~\cite{yang2024deformable, cai2024gic} to minimize the L1 and Structural Similarity Index Measure (SSIM) losses between rendered frames $\hat{\Iv}_t$ and corresponding ground-truth observations $\Iv_t$.
We also regard all Gaussian kernels as isotropic and apply L1 regularization on scales for simplicity and reconstruction quality~\cite{zhong2025reconstruction, yugay2023gaussian, chen2023neusg}.
The overall objective function is formulated as
\begin{align}
\label{eq:obj_gaus}
    \min_{\Gc^*,\Dc}\left[\Lc_1(\hat\Iv_t, \Iv_t) + \lambda_\text{SSIM}\Lc_\text{SSIM}(\hat\Iv_t, \Iv_t) + \lambda_s\norm{s_t}_1\right],
\end{align}
with hyperparameters $\lambda_\text{SSIM}$ and $\lambda_s$.

\subsection{Continuum Trajectory Estimation}
\label{subsec:trajectory}
To bridge the gap between unevenly distributed Gaussian particles and the corresponding solid continuum, we follow~\cite{cai2024gic} to fill internal volumes, forming continuum particles $\{\pv_t\}$ with uniform scale $s$ and density $\sigma$. 
Existing methods for material-specific system identification often rely on sparse visual~\cite{cai2024gic, li2023pac, cao2025neuma} or geometrical~\cite{cai2024gic} cues along $T$ frames to optimize a small set of parameters. 
However, since we do not assume a known constitutive backbone or material types, these sparse signals may lead to unconstrained behaviors at unobserved time steps or even cause training instability. 
Inspired by system identification~\cite{ma2023learning, herde2025poseidon, kang2024pig} using state data per simulation step,
we propose leveraging particle trajectories over $N$ simulation steps as dense geometrical supervision to better constrain the simulated behaviors of continuum particles. 

Given that the deformation network trained in~\cref{subsec:gaussian} can produce rough trajectories for dynamic Gaussians, we fine-tune it to establish trajectories for continuum particles.
Specifically, we initialize the canonical space of particles $\xv_0$ with the filled continuum $\pv_1$ at $t=1$ and apply deformations similar to~\cref{eq:deform_gaus},
\begin{align}
\label{eq:deform_sparse}
    \xv_t=\xv^*+\wv(\xv^*,t)\Bv^{\xv}(t),\;\forall t=0,1,\dots,T,
\end{align}
resulting in per-frame deformed particles $\{\xv_{t}\}$.
Then we optimize this deformation with all filled continuums $\{\pv_t\}$ using the Chamfer Distance ~\cite{erler2020points2surf,ma2020neural}
\begin{align}
    \min_{\xv^*,\Dc'} \left[\Lc_\text{CD}(\{\xv_{t}\}_{t=0}^T, \{\pv_t\}_{t=0}^T)\right],
\end{align}
where $\Dc'$ denotes the fine-tuned deformation network.
Unlike the deformation network $\Dc$ trained in~\cref{subsec:gaussian}, this fine-tuned version $\Dc'$ is aware of the internal particles absence in Gaussians.
Further, the learned motion basis with temporal positional embeddings enables it for temporal interpolation, allowing us to probe the motion between observed discrete time steps $t=0,1,\dots,T$.
Formally, we have particle positions at each simulation time step
\begin{align}
\label{eq:deform_dense}
    \xv_t=\xv^*+\wv(\xv^*,t)\Bv^{\xv}(t),\;\forall t=0,\tau,\dots,N\tau,
\end{align}
where $\tau$ is the time interval between two simulation steps. We then consider these temporally dense particle trajectories as pseudo ground truth for later optimization.

\subsection{Material-Agnostic System Identification}
\label{subsec:system}
Inspired by NCLaw~\cite{ma2023learning}, we parameterize the elastic and plastic constitutive laws using neural networks, denoted as $\Ec_{\theta^e}$ and $\Pc_{\theta^p}$, where $\theta^e$ and $\theta^p$ are their respective network parameters.
Both networks share the same multi-layer perceptron (MLP) architecture with a single hidden layer.
Further, they incorporate two physical priors, frame indifference and undeformed state equilibrium, enforced by using rotation-invariant input representations and eliminating bias terms, respectively.
While we adopt the same parameterization for material-agnostic modeling, we introduce a more challenging setting by relaxing the input requirements from dense, complete state sequences $\{\sv_t\}_{n=0}^N$ to sparse visual observations $\{\Iv_t\}_{t=0}^T$.
To this end, we follow an analysis-by-synthesis scheme.
For the initial particle state $\sv_0$, we use positions $\xv_0$ derived from~\cref{eq:deform_dense}, an optimizable velocity $\vv_0$, and identity deformation gradients $\Fv_0$.
After we evolve particle positions $\{\hat{\xv}_t\}_{t=\tau}^{N\tau}$ through time integration as described in~\cref{eq:transfer}, we optimize $\theta^e$ and $\theta^p$ with
\begin{align}
\label{eq:obj_constitutive}
    \min_{\theta^e,\theta^p}\left(\Lc_\text{geo} + \Lc_\text{sil}\right),
\end{align}
where $\Lc_\text{geo}$ supervises particle positions $\hat{\xv}_t$ with estimated trajectories $\xv_t$ and $\Lc_\text{sil}$ compares rendered masks $\hat{\Mv}_t$ with object silhouettes $\Mv_t$ extracted from frame $\Iv_t$
\begin{align}
\label{eq:loss_geo}
    \Lc_\text{geo} &= \Lc_\text{traj} = \Lc_1\left(\{\hat{\xv}_t\}_{t=0}^{N\tau},\{\xv_t\}_{t=0}^{N\tau}\right), \\
    \Lc_\text{sil} &= \Lc_1\left(\{\hat{\Mv}_t\}_{t=0}^{N\tau},\{\Mv_t\}_{t=0}^{N\tau}\right).
\end{align}

While not assuming known material types in our problem setting, we note that \model still requires a pre-trained constitutive model from~\cite{ma2023learning} as a stable initialization.

%% file: sections/4_experiments.tex
\section{Experiments}
\subsection{Experimental Settings}
\minisection{Datasets.}
We conduct our experiments on the PAC-NeRF dataset~\cite{li2023pac} and the Spring-Gaus dataset~\cite{zhong2025reconstruction}.
The PAC-NeRF dataset consists of synthetic objects simulated using the Material Point Method (MPM) and includes a variety of material types such as elastic bodies, plasticine, sand, and Newtonian and non-Newtonian fluids.
This dataset provides 45 multi-view RGB video sequences that allow for the estimation of both object geometry and physical properties and approximately 14 frames per viewpoint.
The Spring-Gaus dataset contains two subsets, namely, Synthetic and Real.
The Synthetic subset focuses on the reconstruction and simulation of elastic objects, containing 30 frames in each of 10 viewpoints.
The Real subset provides both dense-view static and sparse-view dynamic captures of elastic objects.
We evaluate the performance of observable state simulation using both datasets, while the Spring-Gaus dataset is additionally utilized for future state prediction. To obtain object masks, we apply off-the-shelf matting with~\cite{lin2021real,kirillov2023segment}.

\minisection{Baselines.}
To assess the performance of our approach, we compare it against four state-of-the-art methods for system identification.
PAC-NeRF~\cite{li2023pac} combines neural radiance fields with differentiable physics to jointly infer geometry and material properties.
Spring-Gaus~\cite{zhong2025reconstruction} integrates a spring-mass system into Gaussian representations for efficient elastic object simulation.
NeuMA~\cite{cao2025neuma} fine-tunes a neural adapter on expert-designed physical models to better align reconstructed dynamics with visual observations.
GIC~\cite{cai2024gic} leverages a two-stage pipeline to obtain the geometry representation and physical properties sequentially.
These baselines represent the most relevant prior work and provide a strong comparison for evaluating our method.
We note that they all necessitate material priors.
To be concrete, PAC-NeRF and GIC roughly categorize materials into five groups and adopt a certain set of physical models for each group.
Spring-Gaus generally focuses on elastic objects, thereby being material-specific.
And NeuMA leverages a scene-specific constitutive model as the material prior.

\minisection{Metrics.}
We evaluate our method using several standard metrics to assess both geometric reconstruction quality and physical property estimation accuracy.
Metrics including 1) Chamfer Distance (CD)~\cite{erler2020points2surf,ma2020neural} with units expressed in $10^3\:\text{mm}^2$; 2) Peak Signal-to-Noise Ratio (PSNR)~\cite{hore2010image}; 3) Structural Similarity Index Measure (SSIM)~\cite{wang2004image}.
The CD quantifies the similarity between two point distributions, while PSNR and SSIM assess the visual quality.

\newcolumntype{a}{>{\columncolor{Gray}}c}
\begin{table*}[t]
\centering
\footnotesize
\setlength{\tabcolsep}{0.4em}
\begin{adjustbox}{width=\linewidth}
\begin{tabular}{ll|ccccca|ccccccca}
    \toprule
     \multicolumn{2}{c|}{} & \multicolumn{6}{c|}{PAC-NeRF} & \multicolumn{8}{c}{Spring-Gaus} \\
    & Method & Newtonian & Non-Newtonian & Elasticity & Plasticine & Sand & Mean & Torus & Cross & Cream & Apple & Paste & Chess & Banana & Mean \\
    \midrule
    & PAC-NeRF~\cite{li2023pac} & 0.277 & 0.236 & 0.238 & 0.429 & \textbf{0.212} & 0.278   & 4.92 & 1.10 & 0.77 & 1.11 & 3.14 & 0.96 & 2.77 & 2.11 \\
    & Spring-Gaus~\cite{zhong2025reconstruction} & - & - & - & - & - & - & 0.17 & 0.48 & 0.36 & 0.38 & 0.19 & 1.80 & 2.60 & 0.85 \\ 
    & NeuMA~\cite{cao2025neuma} & - & - & - & - & - & - & 4.59 & \textbf{0.06} & \textbf{0.08} & \textbf{0.04} & 0.71 & \textbf{0.05} & \textbf{0.03} & 0.79 \\ 
    & GIC~\cite{cai2024gic} & 0.243 & \textbf{0.195} & \textbf{0.178} & \textbf{0.196} & 0.250 & 0.212 & 0.13 & 0.13 & 0.14 & 0.15 & 0.17 & 0.41 & \textbf{0.03} & 0.17 \\ 
    \rowcolor{Gray}\cellcolor{white}\multirow{-5}{*}{\rotatebox{90}{CD~$\downarrow$}}& \model (Ours) & \textbf{0.233} & 0.198 & 0.192 & 0.201 & 0.229 & \textbf{0.210} & \textbf{0.08} & 0.10 & 0.19 & 0.16 & \textbf{0.13} & 0.19 & 0.08 & \textbf{0.13} \\
    \bottomrule
\end{tabular}
\end{adjustbox}
\caption{\textbf{Observable state simulation on PAC-NeRF and Spring-Gaus Synthetic datasets.} Despite being the only technique that does not use expert-designed physical models (noted by the gray row), \model performs better than or on par with existing baselines.}
\label{tab:observable}
\vspace{-1.0em}
\end{table*}

\begin{table}[t]
\centering
\footnotesize
\setlength{\tabcolsep}{0.4em}
\adjustbox{width=\linewidth}{
\begin{tabular}{ll|ccccca}
\toprule
    & Method & Bun & Burger & Dog & Pig & Potato & Mean \\
\midrule
    & Spring-Gaus~\cite{zhong2025reconstruction}& 30.69 & 34.01 & 32.10 & 34.97 & 32.72 & 32.90 \\
    & NeuMA~\cite{cao2025neuma} & 31.27 & 23.78 & 25.61 & 25.40 & - & 26.51 \\
    & GIC~\cite{cai2024gic} & 34.68 & 40.45 & 37.17 & 38.32 & 39.95 & 38.11 \\
    \rowcolor{Gray}\cellcolor{white}\multirow{-4}{*}{\rotatebox{90}{PSNR~$\uparrow$}} & Ours & \textbf{40.26} & \textbf{42.12} & \textbf{38.97} & \textbf{40.84} & \textbf{43.40} & \textbf{41.12} \\
\midrule
    & Spring-Gaus & 0.992 & 0.994 & 0.994 & 0.996 & 0.992 & 0.994 \\
    & NeuMA & 0.994 & 0.993 & 0.995 & 0.995 & - & 0.994 \\
    & GIC & 0.995 & \textbf{0.997} & \textbf{0.997} & 0.997 & 0.996 & 0.996 \\
    \rowcolor{Gray}\cellcolor{white}\multirow{-4}{*}{\rotatebox{90}{SSIM~$\uparrow$}} & Ours & \textbf{0.998} & \textbf{0.997} & \textbf{0.997} & \textbf{0.998} & \textbf{0.998} & \textbf{0.997} \\
\bottomrule
\end{tabular}}
\caption{\textbf{Observable state simulation on Spring-Gaus Real dataset.} \model achieves state-of-the-art performance.}
\label{tab:spring_gaus_real}
\vspace{-1.0em}
\end{table}

\minisection{Implementation details.}
In dynamic reconstruction, we follow~\cite{cai2024gic} to optimize canonical Gaussians $\Gc^*$ and the deformation network $\Dc$ for 40K iterations. Then the deformation network is fine-tuned to estimate particle trajectories from filled continuum sequences for 10K iterations. In system identification, we set the simulation time interval $\tau$ to $1/200$ the frame duration. We adopt the RAdam~\cite{liu2019variance} optimizer to optimize the initial velocity $\vv_0$ for 100 steps and constitutive model parameters $\theta^e, \theta^p$ for 1000 steps. All experiments are conducted on a single NVIDIA A100 GPU.

\subsection{Comparison Results}
\minisection{Observable state simulation.}
In~\cref{tab:observable}, we compare \model with existing vision-based system identification methods for observable state simulation.
The results indicate that \model outperforms all other methods in terms of CD, highlighting its effectiveness in reconstructing object dynamics across diverse materials. 
Further, \cref{tab:spring_gaus_real} evaluates on real data.
Thanks to the more flexible constitutive representation, \model achieves state-of-the-art results.
This demonstrates the potential of \model as an adaptive and generalizable approach for system identification.

\begin{figure*}[t]
    \centering
    \includegraphics[width=\linewidth]{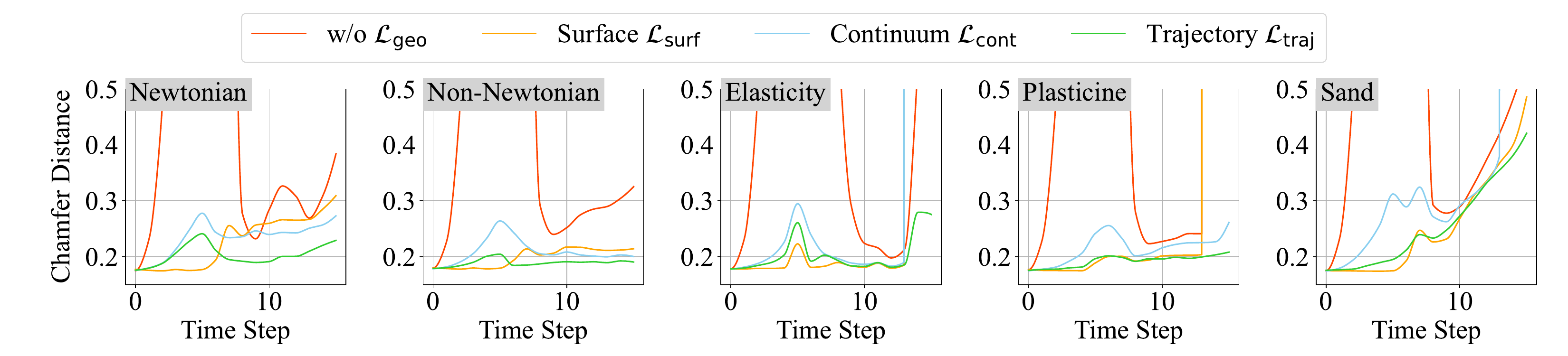}
    \vspace{-1.0em}
    \caption{\textbf{Ablation study of geometrical objectives $\Lc_\text{geo}$ on PAC-NeRF dataset.} We compare three types of geometrical objectives against a baseline without geometrical guidance and plot their simulation errors using Chamfer Distance (CD). Overall, dense trajectory guidance achieves the lowest and most stable errors. The figure is best viewed in color.}
    \vspace{-1.0em}
    \label{fig:ablation_supervision}
\end{figure*}

\minisection {Future state simulation.}
In~\cref{tab:spring_gaus_future}, we provide a detailed evaluation of future state simulation performance.
The results indicate that \model consistently outperforms all other methods while slightly underperforming GIC.
Since NeuMA~\cite{cao2025neuma} does not originally experiment on the synthetic subset of Spring-Gaus, we report our reproduction results.
To be concrete, we use the closest pre-trained model, Jelly, from NCLaw~\cite{ma2023learning} as the material prior and conduct LoRA~\cite{hu2022lora} fine-tuning with hyperparameters from its original implementation.
However, we observe unstable optimization processes for most categories.
This is probably rooted in its sparse pixel supervision, which only operates once per frame.
With dense geometrical constraints supervising hundreds of times per frame, \model achieves significantly better results on both geometrical accuracy and rendering quality.
This suggests that \model effectively captures the underlying dynamics governing future states, allowing it to generate predictions that remain structurally and visually consistent over time.
Unlike baseline approaches that rely on predefined material priors, \model demonstrates the ability to infer relevant properties directly from visual observations, making it more adaptable to a wide range of scenarios.
On the other hand, while dense trajectory guidance enables overfitting to observations, the lack of category-level regularization, unlike hand-crafted constitutive models, limits its generalization under scarce data.
This leads to inferior performance than GIC, which can benefit from its pre-known constitutive model as an oracle.
More analysis on the enhanced generalization ability of \model with more data is in~\cref{sec:supp:quantitative} of \supp

\begin{table}[t]
\centering
\footnotesize
\setlength{\tabcolsep}{0.4em}
\adjustbox{width=\linewidth}{
\begin{tabular}{ll|ccccccca}
\toprule
    & Method & Torus & Cross & Cream & Apple & Paste & Chess & Banana & Mean \\
\midrule
    & PAC-NeRF~\cite{li2023pac}  & 2.47 & 3.87 & 2.21 & 4.69 & 37.70 & 8.20 & 66.43 & 17.94 \\
    & Spring-Gaus~\cite{zhong2025reconstruction}  & 2.38 & 1.57 & 2.22 & 1.87 & 7.03 & 2.59 & 18.48 & 5.16 \\
    & NeuMA~\cite{cao2025neuma} & 1.17 & 0.40 & 1147.4 & \textbf{0.05} & 44.26 & \textbf{0.33} & 0.30 & 170.56 \\
    & GIC~\cite{cai2024gic}  & 0.75 & 1.09 & \textbf{0.94} & 0.22 & 2.79 & 0.77 & \textbf{0.12} & \textbf{0.95} \\
    \rowcolor{Gray}\cellcolor{white}\multirow{-5}{*}{\rotatebox{90}{CD~$\downarrow$}} & \model (Ours)  & \textbf{0.16} & \textbf{0.38} & 1.51 & 2.64 & \textbf{1.75} & 1.70 & 0.24 & 1.20 \\
\midrule
    & PAC-NeRF~\cite{li2023pac}  & 17.46 & 14.15 & 15.37 & 19.94 & 12.32 & 15.08 & 16.04 & 15.77 \\
    & Spring-Gaus~\cite{zhong2025reconstruction}  & 16.83 & 16.93 & 15.42 & 21.55 & 14.71 & 16.08 & 17.89 & 17.06 \\ 

    & NeuMA~\cite{cao2025neuma} &16.61 & 19.67 &2.32 & 26.25 &13.42 & \textbf{20.85} & 23.16 & 17.47\\
    & GIC~\cite{cai2024gic}  & 20.24 & 30.51 & \textbf{19.15} & \textbf{26.89} & \textbf{16.31} & 18.44 & \textbf{29.29} & \textbf{22.98} \\
    \rowcolor{Gray}\cellcolor{white}\multirow{-5}{*}{\rotatebox{90}{PSNR~$\uparrow$}}& \model (Ours)  & \textbf{22.42} & \textbf{33.73} & 16.45 & 21.19 & 15.73 & 17.28 & 26.56 & 21.91 \\
\midrule
    & PAC-NeRF~\cite{li2023pac}  & 0.919 & 0.906 & 0.858 & 0.878 & 0.819 & 0.848 & 0.886 & 0.870 \\
    & Spring-Gaus~\cite{zhong2025reconstruction}  & 0.919 & 0.940 & 0.862 & 0.902 & 0.872 & 0.881 & 0.904 & 0.897 \\
    & NeuMA~\cite{cao2025neuma} & 0.942 & 0.948 & 0.889 & \textbf{0.964} & 0.889 & \textbf{0.933} & 0.964 & \textbf{0.933} \\
    & GIC~\cite{cai2024gic}  & 0.942 & 0.939 & \textbf{0.909} & 0.948 & \textbf{0.894} & 0.912 & 0.964 & 0.930 \\
    \rowcolor{Gray}\cellcolor{white}\multirow{-5}{*}{\rotatebox{90}{SSIM~$\uparrow$}}& \model (Ours)  & \textbf{0.955} & \textbf{0.967} & 0.852 & 0.898 & 0.891 & 0.878 & \textbf{0.965} & 0.915 \\
\bottomrule
\end{tabular}}
\caption{\textbf{Future state simulation on Spring-Gaus Synthetic dataset.} \model outperforms all baselines except for GIC.}
\label{tab:spring_gaus_future}
\vspace{-1.0em}
\end{table}

\begin{figure}[t]
    \centering
    \includegraphics[width=\linewidth]{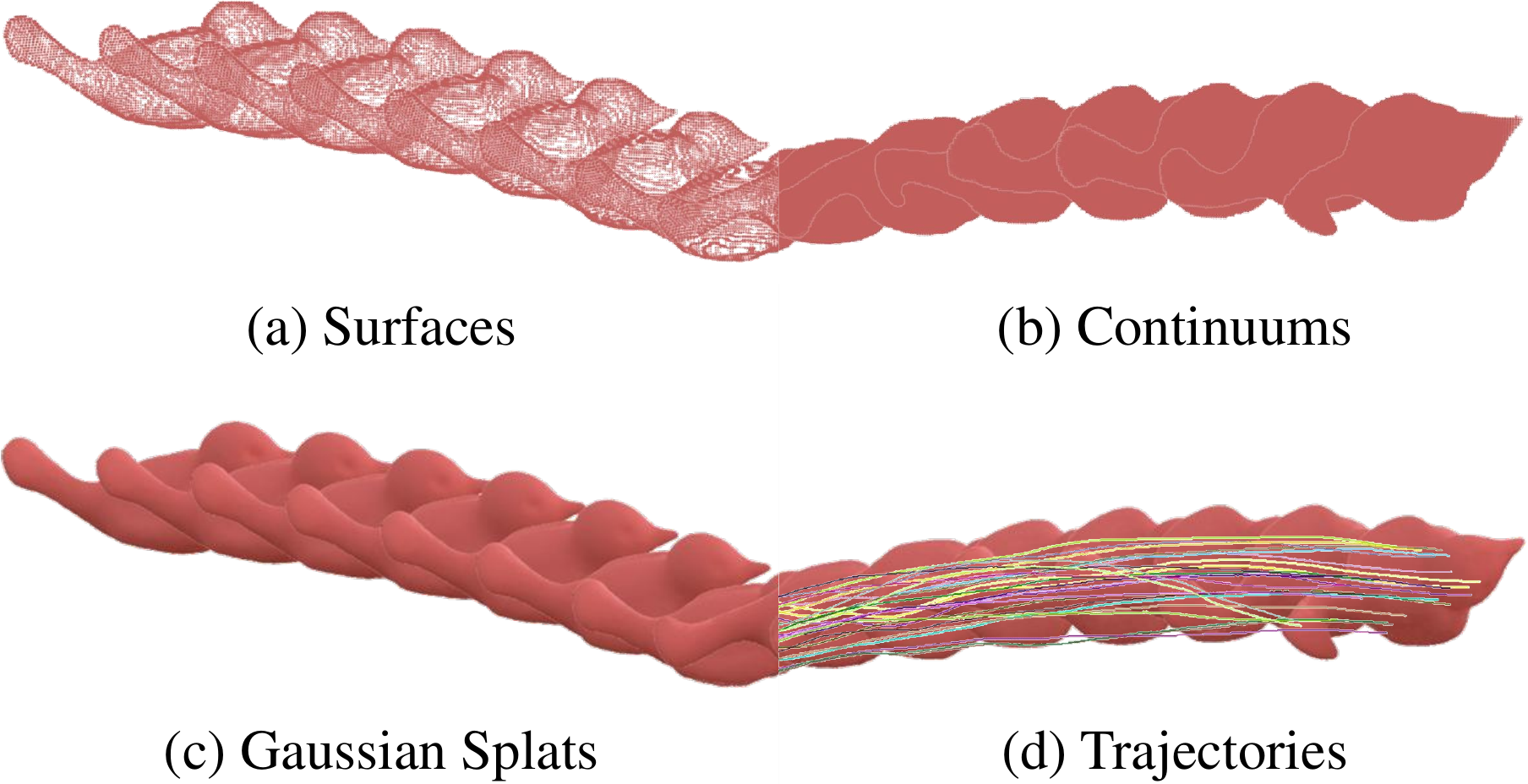}
    \vspace{-2.0em}
    \caption{\textbf{Visualization of geometrical objectives $\Lc_\text{geo}$ on PAC-NeRF Bird.} The fine-tuned deformation network $\Dc'$ captures smooth and accurate motion for continuums, including low-frequency translation of the body and high-frequency swing of the tail. We space adjacent frames in the horizontal axis for clarity.}
    \label{fig:visualization_supervision}
    \vspace{-0.5em}
\end{figure}

\subsection{Ablation Study}
We investigate the impact of geometrical objectives here, while referring the readers to more ablation studies in~\cref{sec:supp:ablation} of \supp
\cref{fig:ablation_supervision} validates the effectiveness of dense trajectory supervision.  
As shown in~\cref{fig:visualization_supervision}, we evaluate three geometrical objectives by replacing the $\Lc_\text{geo}$ term in optimizing the neural constitutive model objective,  \cref{eq:obj_constitutive}. We define them formally as follows.
(1) \textbf{Surface}: per-frame surface alignment between the simulated continuum and the filled continuum,
\begin{align}
    \Lc_\text{surf} = \Lc_\text{CD}\left(\left\{\Sc\left(\hat{\xv}_t\right)\right\}_{t=0}^T, \left\{\Sc\left(\pv_t\right)\right\}_{t=0}^T\right),
\end{align}
where $\Sc$ denotes the surface extraction by thresholding Gaussian opacities~\cite{cai2024gic}. (2) \textbf{Continuum}: similar to Surface while between solid continuums instead of surfaces,
\begin{align}
    \Lc_\text{cont} = \Lc_\text{CD}\left(\left\{\hat{\xv}_t\right\}_{t=0}^T, \left\{\pv_t\right\}_{t=0}^T\right).
\end{align}
(3) \textbf{Trajectory}: per-simulation-step trajectory alignment between the simulated continuum $\{\hat{\xv}_t\}_{t=0}^{N\tau}$ and the continuum predicted by the deformation network $\{\xv_t\}_{t=0}^{N\tau}$, as defined in~\cref{eq:loss_geo}.  
By analyzing per-frame errors, we first confirm the necessity of geometrical guidance, as the baseline (red curve) exhibits significantly higher CD errors across all material types.
Surface-based supervision (yellow curve) shows moderate error reduction but remains less stable over time, especially in Plasticine and Sand.
Continuum-based supervision (blue curve) further reduces errors but struggles with certain materials, e.g., Sand.
These error overshoots are probably due to supervising relatively complex neural constitutive models with sparse signals, which might lead to overfitting since the simulated behaviors between observed discrete time steps are not well constrained.
Finally, by enforcing dense alignment across simulation steps, trajectory-based supervision (green curve) achieves the lowest and most stable errors across all materials.

\begin{figure}[t]
    \centering
    \includegraphics[width=\linewidth]{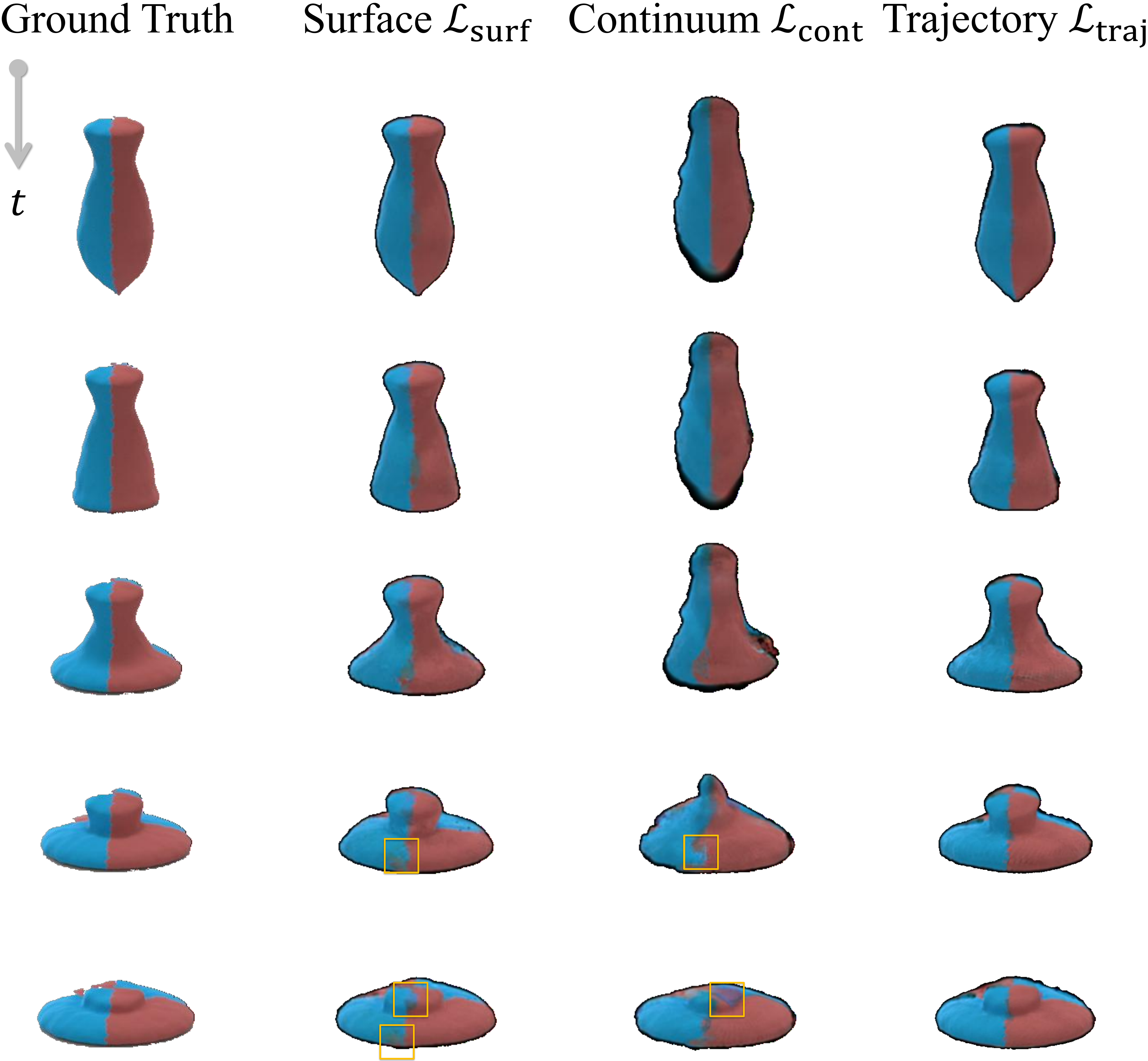}
    \caption{\textbf{Qualitative comparison of geometrical objectives $\Lc_\text{geo}$ on PAC-NeRF Cream.} Using sparse surfaces or solid continuum as geometrical cues can lead to physically implausible deformations, while trajectory guidance mitigates these issues by introducing temporally denser constraints. Better viewed when zoomed in.}
    \label{fig:qualitative_reconstruction}
    \vspace{-1.0em}
\end{figure}

\subsection{Qualitative Analysis}
\minisection{Dynamics reconstruction.}
We visualize the reconstructed Cream object by using the alternative objectives in~\cref{fig:qualitative_reconstruction}.
Here, we observe physically implausible deformations in results for surface and continuum supervision.
Under uniform force, the object exhibits inconsistent distortion, especially in the boundary between blue stripes and red stripes.
We attribute this to two main reasons.
1) Chamfer Distance based geometrical constraint focuses mainly on the overall structure.
While it encourages alignment between simulated and reference shapes, they do not enforce point-wise correspondence.
A single particle can thus be matched with different ground truth particles in different time steps.
As a result, local mismatches can accumulate, leading to inaccurate deformations.
2) Sparse supervision fails to constrain intermediate motion dynamics.  
Surface and continuum objectives provide alignment only at discrete time steps, leaving large gaps in supervision.  
Without explicit constraints on how the object deforms over time, the neural constitutive model may interpolate arbitrarily, leading to unstable and unrealistic motion patterns.

\begin{figure}
    \centering
    \includegraphics[width=\linewidth]{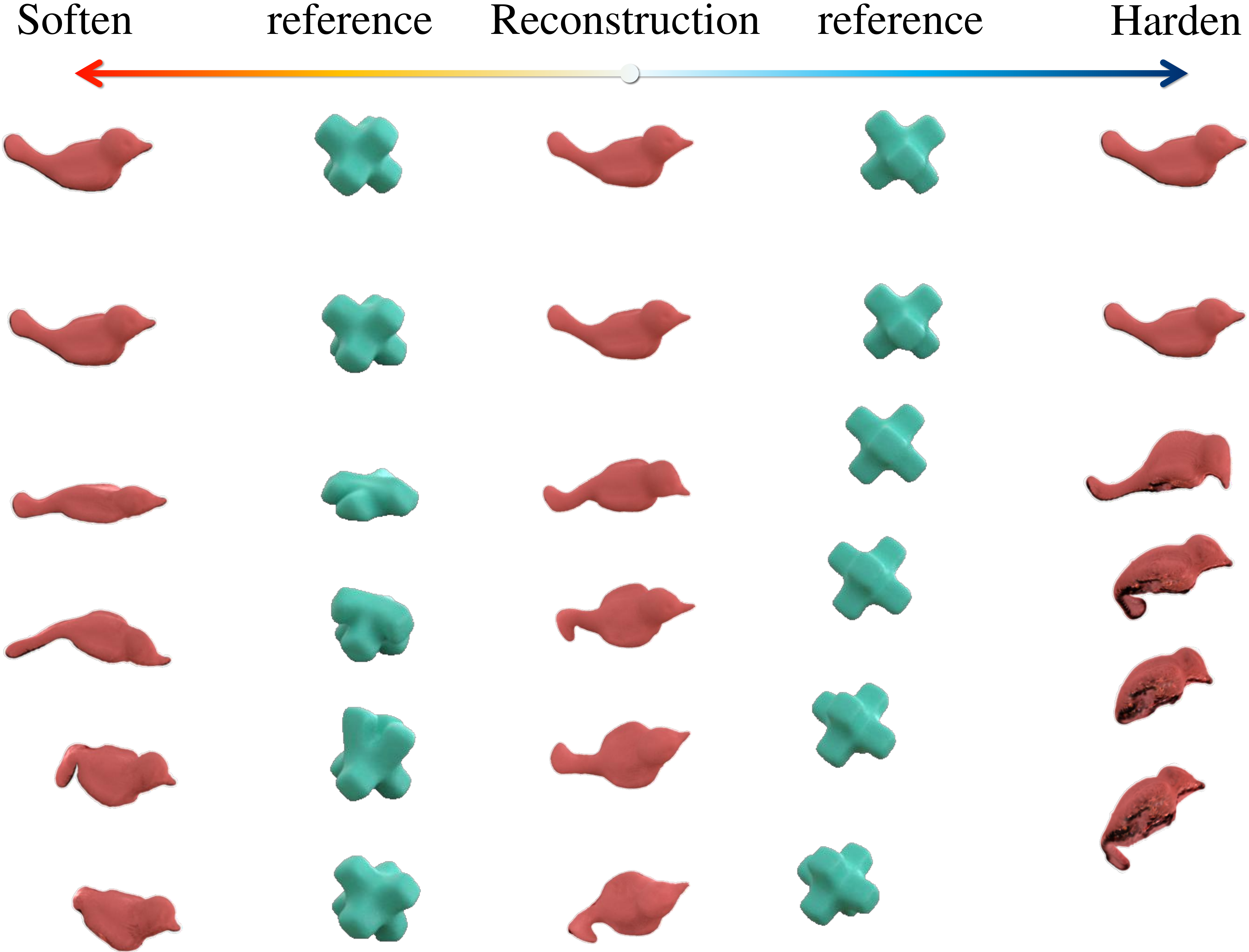}
    \caption{\textbf{Intra-class generalization.} We first optimize the geometry for PAC-NeRF Bird and then use the constitutive models optimized for reference elastic objects to predict novel dynamic behaviors, thereby softening or hardening the Bird respectively.}
    \label{fig:qualitative_intra}
\end{figure}

\begin{figure}
    \centering
    \includegraphics[width=\linewidth]{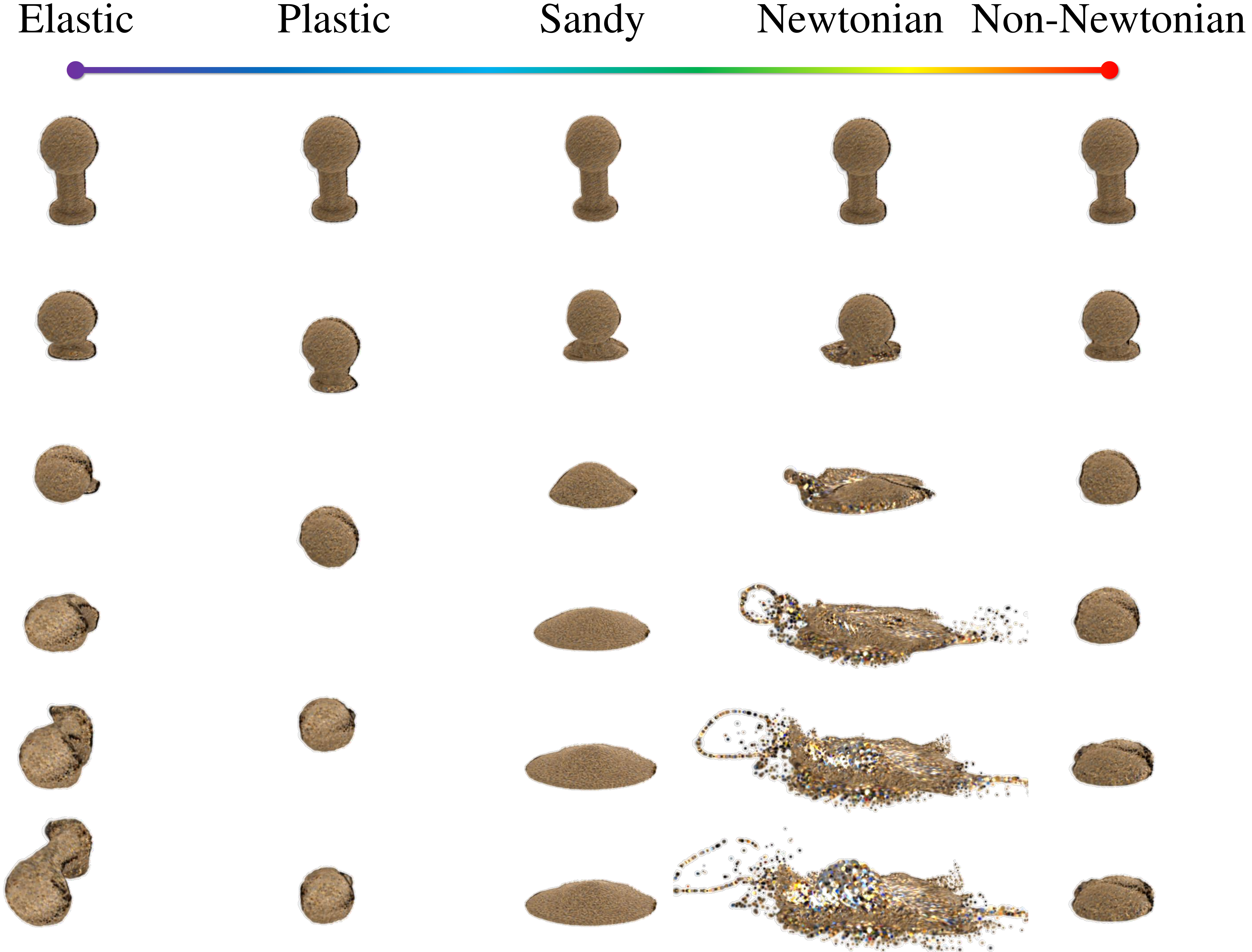}
    \caption{\textbf{Inter-class generalization.} We train multiple neural constitutive models with different materials and apply them to the PAC-NeRF Trophy. While sharing the same initial state, the Trophies with different materials exhibit distinct behaviors over time.}
    \label{fig:qualitative_inter}
\end{figure}

\minisection{Dynamics generalization.}
We further study the generalization ability for \model.
\cref{fig:qualitative_intra} examines how \model can be used to modify the dynamic characteristics of a single object (PAC-NeRF Bird).
Specifically, we first reconstruct the geometrical representation of the Bird and optimize neural constitutive models for two elastic cross-shaped objects.
These optimized constitutive models are then used as references to drive the simulation of the Bird, which deforms accordingly, exhibiting fluid-like softness on the left and increased rigidity and bounce on the right.
Additionally, we see continuous transitions between soft and rigid behaviors when interpolating the neural constitutive model weights.
In~\cref{fig:qualitative_inter}, we extend this study to inter-class generalization by applying neural constitutive models trained on different materials to a novel object, the PAC-NeRF Trophy.
The results show that despite sharing the same initial state, each material model induces distinct deformation patterns.
These observations confirm that \model effectively captures generalizable intrinsic dynamics.
More qualitative results can be found in~\cref{sec:supp:qualitative} of \supp

%% file: sections/5_conclusion.tex
\subsection{Discussion}
While \model excels in material-agnostic system identification, several limitations remain.
First, it relies on multi-view video input, restricting applicability to single-view or sparse observations.
Second, our approach assumes that internal geometries and deformations can be inferred from surface motion, following a spatially coherent model.
This assumption may break down in materials with complex internal flows, such as granular media or multiphase fluids.
Finally, \model does not explicitly disentangle material properties from external forces in system identification, assuming a uniform gravity field for all objects.
Future works aim to relax these assumptions for broader applicability.

\section{Conclusion}
We present \model, a material-agnostic system identification framework.
By integrating neural constitutive models with dense geometry guidance, \model effectively captures object dynamics across diverse materials.
Our approach reconstructs dynamic Gaussians from multi-view videos and fine-tunes a deformation model within a solid continuum.
This enables us to extract dense motion cues, which, combined with visual observations, guide the learning of neural constitutive models.
Through this formulation, \model generalizes beyond traditional system identification methods that rely on pre-defined material priors.
Extensive experiments highlight its ability to reconstruct observable dynamics and to generalize for novel interactions.

%% file: sections/6_acknowledgment.tex
\section*{Acknowledgment}
This work was supported in part by U.S. NSF grant DBI-2238093, and in part by the Institute of Information \& Communications Technology Planning \& Evaluation (IITP) grant funded by the Korean Government (MSIT) (No. RS-2024-00457882, National AI Research Lab Project).

%% file: sections/X_suppl.tex
\clearpage
\setcounter{page}{1}
\setcounter{section}{0}
\renewcommand{\thesection}{\Alph{section}}
\maketitlesupplementary

\section{Background}
\subsection{3D Gaussian Splatting (3DGS)}
3DGS~\cite{kerbl20233d} represents a scene using a set of 3D Gaussian kernels $\Gc=\{\muv(i), \Sigmav(i), \cv(i), \sigma(i)\}$, parameterized with center coordinates $\muv(i) \in \Rb^3$, covariance matrix $\Sigmav(i) \in \Rb^{3 \times 3}$ defining the shape and orientation of the Gaussian, color attribute $\cv(i)$, and the opacity $\sigma(i)$. 
Each point is denoted as
\begin{equation}
    \Gc(\xv) = \text{exp}(-\frac{1}{2} (\xv - \muv(i))^T \Sigmav(i)^{-1} (\xv - \muv(i))),
\end{equation}
where the covariance matrix can be factorized into rotation and scaling components as $\Sigmav(i) = \Rv(i) \Sv(i) \Sv(i)^T \Rv(i)^T$.
A rotation matrix $\Rv(i)$ is represented by a quaternion vector $\qv(i) \in \Rb^4$, and a diagonal scaling matrix $\Sv(i)$ is characterized by $\sv(i) \in \Rb^3$ which defines the anisotropic spread of the Gaussian.
During rendering, each 3D Gaussian is projected onto the 2D image plane, and an image is rendered via point-based alpha blending, where the rendered color $\Iv$ and the corresponding foreground mask $\Mv$ at pixel $\uv$ are
\begin{align}
    \Iv(\uv) &= \sum_{i=1}^{G} T(i)\sigma(i) \cv(i), \\
    \Mv(\uv) &= \sum_{i=1}^{G} T(i) \sigma(i).
\end{align}
Here, $T(i) =\prod_{j=1}^{i-1} (1 - \sigma(j))$ is the accumulated transmittance at point $i$, and $G$ is the number of ordered Gaussians.

\subsection{Material Point Method (MPM)}
\label{subsec:supp:mpm}
MPM~\cite{jiang2016material,stomakhin2013material} is a hybrid Eulerian-Lagrangian approach used for simulating materials with complex dynamics, including elastic and plastic deformations. MPM represents the simulated object using a set of discrete particles, which store material properties such as mass, velocity, and deformation gradients. These particles interact with a background grid where forces are computed, and the governing equations of motion are solved. The method consists of four key steps: particle-to-grid transfer, solving the governing equations on the grid to do the grid update, grid-to-particle transfer, and updating particle states.

\begin{enumerate}[itemsep=5pt, parsep=5pt]
    \item \textbf{Initialization.}
    
    Each material point \( i= 1,2, ..., Q \), where \( Q \) is the total number of material points, at every time step \( t \) is initialized by defining its position \( \xv_i^t \), velocity \( \vv_i^t \), deformation gradient \( \Fv_i^t \), and affine velocity field \( \Cv_i^t \). The initial values are set as \( \xv_i^0 = \xv_i, \vv_i^0 = \mathbf{0}, \Fv_i^0 = \Iv, \Cv_i^0 = \mathbf{0} \). And for each grid node \( g \in \Gc \) at every time step \( t \), we define its grid mass \( m_g^t \) and grid velocity \( \vv_g^t \). The initial values are set as \( m_g^0 = 0, \vv_g^0 = \mathbf{0} \).
    
   \item \textbf{Stress update.}
   
    The stress state of each material point is determined by its constitutive model. The first Piola-Kirchhoff stress tensor is computed as:
    \begin{equation}
        \Pv_i^t = \frac{\partial \Psi_i}{\partial \Fv_i^t}, \quad \forall i \in Q.
    \end{equation}
    where \( \Pv_i^t \) represents the stress tensor, and \( \Psi_i \) is the strain energy function describing the material's response to deformation.

    \item \textbf{Particle-to-grid transfer.}
    
    Mass and momentum are transferred between particles and grid nodes, according to the interpolation functions. We denote the interpolation weight of particle \( i \) at grid node \( g \) as \( \omega_{ig}^t \in \mathbb{R} \). The mass contribution at a grid node is computed as:
    \begin{equation}
        m_g^t = \sum_{i \in Q} \omega_{ig}^t m_i, \quad \forall g \in \Gc.
    \end{equation}
    The corresponding momentum at grid nodes is computed as:
    \begin{equation}
        m_g^t \vv_g^t = \sum_{i \in Q} \omega_{ig}^t m_i \left( \vv_i^t + \Cv_i^t (\xv_g - \xv_i^t) \right).
    \end{equation}
    where \( \Cv_i^t = \Bv_i^t (\Dv_i^t)^{-1} \), \( \Bv_i^t \) is a matrix quantity stored at each particle (similar to mass, position, and velocity), and \( \Dv_i^t \) is an inertia-like tensor.\\
    The inertia-like tensor \( \Dv_i^t \) is defined as:
    \begin{equation}
        \Dv_i^t = \sum_{g\in \Gc} \omega_{ig}^t (\xv_g - \xv_i^t)(\xv_g - \xv_i^t)^T,
    \end{equation}
    which accounts for the distribution of particles' positions relative to the particle \( i \). 
    The matrix \( \Bv_i^t \), is given by:
    \begin{equation}
        \Bv_i^t = \sum_{g \in \Gc } \omega_{ig}^t \vv_g^t (\xv_g - \xv_i^t)^T.
    \end{equation}

    \item \textbf{Grid update.}
    
    After transferring mass and momentum, the governing equations of motion are solved to update grid velocities. The velocity at each grid node is updated as:
    \begin{equation}
        \vv_g^{t+1} = \vv_g^t - \frac{\Delta t}{m_g} \sum_{i \in Q} \Pv_i \nabla \omega_{ig}^t V_i + \Delta t\fv.
    \end{equation}
    where \( V_i \) represents the material point volume, and \( \fv \) accounts for external forces such as gravity.
    
    \item \textbf{Grid-to-particle transfer.}
    
    After computing the grid velocities, the updated values are mapped back to the material points. The velocity of a material point is obtained as:
    \begin{equation}
        \vv_i^{t+1} = \sum_{g \in \Gc} \omega_{ig}^t \vv_g^{t+1}.
    \end{equation}
    The new position of each material point is updated using:
    \begin{equation}
        \xv_i^{t+1} = \xv_i^t + \Delta t \vv_i^{t+1}.
    \end{equation}
    To ensure a consistent velocity field, the affine velocity field is updated using:
    \begin{equation}
        \Cv_i^{t+1} = \frac{12}{(b+1) \Delta x^2} \sum_{g \in \Gc} \omega_{ig}^t \vv_g^{t+1} (\xv_g - \xv_i^t)^\top.
    \end{equation}
    where \( b \) is the B-spline degree, and \( \Delta x \) is the grid spacing.

    The velocity gradient is computed as:
    \begin{equation}
        \nabla \vv_i^{t+1} = \sum_{g \in \Gc} \vv_g^{t+1} \nabla {\omega_{ig}^t}^\top.
    \end{equation}
    The trial deformation gradient is then updated as:
    \begin{equation}
        \Fv_{i, \text{trial}}^{t+1} = (\Iv + \Delta t \nabla \vv_i^{t+1}) \Fv_i^t.
    \end{equation}

    \item \textbf{Plasticity correction.}
    
    The trial deformation gradient is modified using a return mapping function to enforce material constraints:
    \begin{equation}
        \Fv_i^{t+1} = \psi_i (\Fv_{i, \text{trial}}^{t+1}), \quad \forall i \in Q.
    \end{equation}
    where \( \psi_i \) ensures that the material satisfies the yield condition.

    \item \textbf{Update particle positions.}
    
    Finally, the updated positions of the material points are determined using:
    \begin{equation}
        \xv_i^{t+1} = \xv_i^t + \Delta t \vv_i^{t+1}.
    \end{equation}
\end{enumerate}

\section{More Ablation Study}
\label{sec:supp:ablation}
\begin{table}[t]
\centering
\footnotesize
\setlength{\tabcolsep}{0.4em}
\begin{adjustbox}{width=\linewidth}
\begin{tabular}{lc|ccccca}
    \toprule
    Pre-train & Fine-tune & Newtonian & Non-Newtonian & Elasticity & Plasticine & Sand & Mean \\
    \midrule
    \multirow{2}{*}{Jelly} & \xmark & 27.271 & 16.079 & 5.469 & 12.519 & 26.119 & 17.491 \\
    & \cmark & 0.233 & 0.198 & 0.192 & 0.201 & 0.229 & 0.210 \\
    \midrule
    \multirow{2}{*}{Plasticine} & \xmark & 2.826 & 1.118 & 24.430 & 1.419 & 2.883 & 6.535 \\
    & \cmark & 0.202 & 0.189 & 4.408 & 0.191 & 0.208 & 1.040 \\
    \midrule
    \multirow{2}{*}{Sand} & \xmark & 62.012 & 0.795 & 2.133 & 6.701 & 0.338 & 14.396 \\
    & \cmark & 43.176 & 0.192 & 0.335 & 1.815 & 0.234 & 9.151 \\
    \bottomrule
\end{tabular}
\end{adjustbox}
\caption{Ablation study of pre-trained models on PAC-NeRF dataset. We report Chamfer Distance (CD) errors for models trained on full state data from~\cite{ma2023learning} and our fine-tuned version. Fine-tuning substantially improves performance, with Jelly pre-training offering the best tradeoffs between different materials.}
\label{tab:ablation_pretrain}
\end{table}

\minisection{Impact of pre-trained models.}
In~\cref{tab:ablation_pretrain}, we initialize the constitutive model parameters, $\theta^e$ and $\theta^p$, of \model using pre-trained checkpoints trained on data with complete state observations, including position, velocity, deformation gradient, and affine momentum at each simulation time step.
The results show that pre-trained models exhibit varying performance across different materials.
Fine-tuning with \model consistently enhances performance across all scenes, demonstrating its robustness to different initializations.  
Unlike system identification with explicit physical parameters, which typically requires scene-specific initial guesses~\cite{li2023pac, cai2024gic}, \model adapts effectively to diverse material behaviors.  
However, for challenging adaptations like Sand $\rightarrow$ Newtonian, large errors persist even after 1000 training steps.  
This limitation, however, opens a door for pre-training neural constitutive models capable of generalizing more effectively across diverse downstream materials.
In our comparison experiments, we adopt the Jelly pre-training for the PAC-NeRF Synthetic dataset and Plasticine for both Spring-Gaus subsets, as the Jelly pre-training can be unstable in long-term roll-outs.
We also observe an interesting phenomenon that the Sand pre-training performs worse than the Plasticine pre-training on Sand objects.
By looking into the ground truth physical parameters of Sand objects, we conjecture that the Plasticine prior provides a smoother initialization due to its visco-plastic nature, which aligns better with the dynamic behavior of the Sand objects used for fine-tuning.
In contrast, the Sand pre-training includes more brittle behaviors not featured in fine-tuning instances.
This suggests that the effectiveness of pre-training may depend on the similarity of dynamic behavior, which cannot be simply measured with Chamfer Distance that focuses on global alignments rather than local motions.

\begin{table}[t]
\centering
\footnotesize
\setlength{\tabcolsep}{0.4em}
\begin{adjustbox}{width=\linewidth}
\begin{tabular}{c|ccccca}
    \toprule
     $\Lc_\text{sil}$ & Newtonian & Non-Newtonian & Elasticity & Plasticine & Sand & Mean \\
    \midrule
    \xmark & 0.410 & 0.412 & 0.238 & 0.500 & 3.071 & 0.926 \\
    \cmark & 0.233 & 0.198 & 0.192 & 0.201 & 0.229 & 0.210 \\
    \bottomrule
\end{tabular}
\end{adjustbox}
\caption{Ablation study of silhouette loss for observable state simulation on PAC-NeRF dataset in terms of Chamfer Distance.}
\label{tab:ablation_silhouette}
\end{table}

\begin{table}[t]
\centering
\footnotesize
\setlength{\tabcolsep}{0.4em}
\begin{adjustbox}{width=\linewidth}
\begin{tabular}{c|ccccca}
    \toprule
     $\Lc_\text{geo}$ & Newtonian & Non-Newtonian & Elasticity & Plasticine & Sand & Mean \\
    \midrule
    \xmark & 0.346 & 0.315 & 0.769 & 8776.226 & 0.519 & 1755.635 \\
    $\Lc_\text{surf}$ & 0.297 & 0.213 & 11500.481 & 6466.605 & 0.443 & 3593.608 \\
    $\Lc_\text{cont}$ & 0.259 & 0.226 & 0.418 & 0.225 & 0.411 & 0.308 \\
    $\Lc_\text{traj}$ & 0.282 & 0.219 & 0.281 & 0.210 & 0.477 & 0.294 \\
    \bottomrule
\end{tabular}
\end{adjustbox}
\caption{Ablation study of geometrical objectives for future state simulation on PAC-NeRF dataset in terms of Chamfer Distance.}
\label{tab:ablation_supervision}
\end{table}

\minisection{Impact of silhouette loss.}
\cref{tab:ablation_silhouette} presents an ablation study evaluating the impact of silhouette loss ($\Lc_\text{sil}$) on observable state simulation in the PAC-NeRF dataset.  
The results show that, when silhouette loss is applied, CD values significantly decrease across all materials.  
This improvement demonstrates that silhouette loss effectively constrains object boundaries, leading to more accurate shape reconstructions and enhanced simulation fidelity.  

\minisection{Impact of different geometrical objectives.}
\cref{tab:ablation_supervision} evaluates the impact of geometrical objectives ($\Lc_\text{geo}$) on future state simulation in the PAC-NeRF dataset.
Without geometrical supervision, the Plasticine category exhibits an extremely high error, indicating unstable and inaccurate predictions.  
Similarly, when using surface-based supervision ($\Lc_\text{surf}$), Elasticity and Plasticine show catastrophic errors, suggesting that relying solely on surface alignment fails to provide robust constraints for highly deformable materials.  
In contrast, continuum-based ($\Lc_\text{cont}$) and trajectory-based ($\Lc_\text{traj}$) supervision significantly reduce errors, with $\Lc_\text{traj}$ achieving the lowest mean CD.
This highlights the crucial role of spatially and temporally dense geometric supervision in training neural constitutive models.

\begin{figure}
    \centering
    \includegraphics[width=\linewidth]{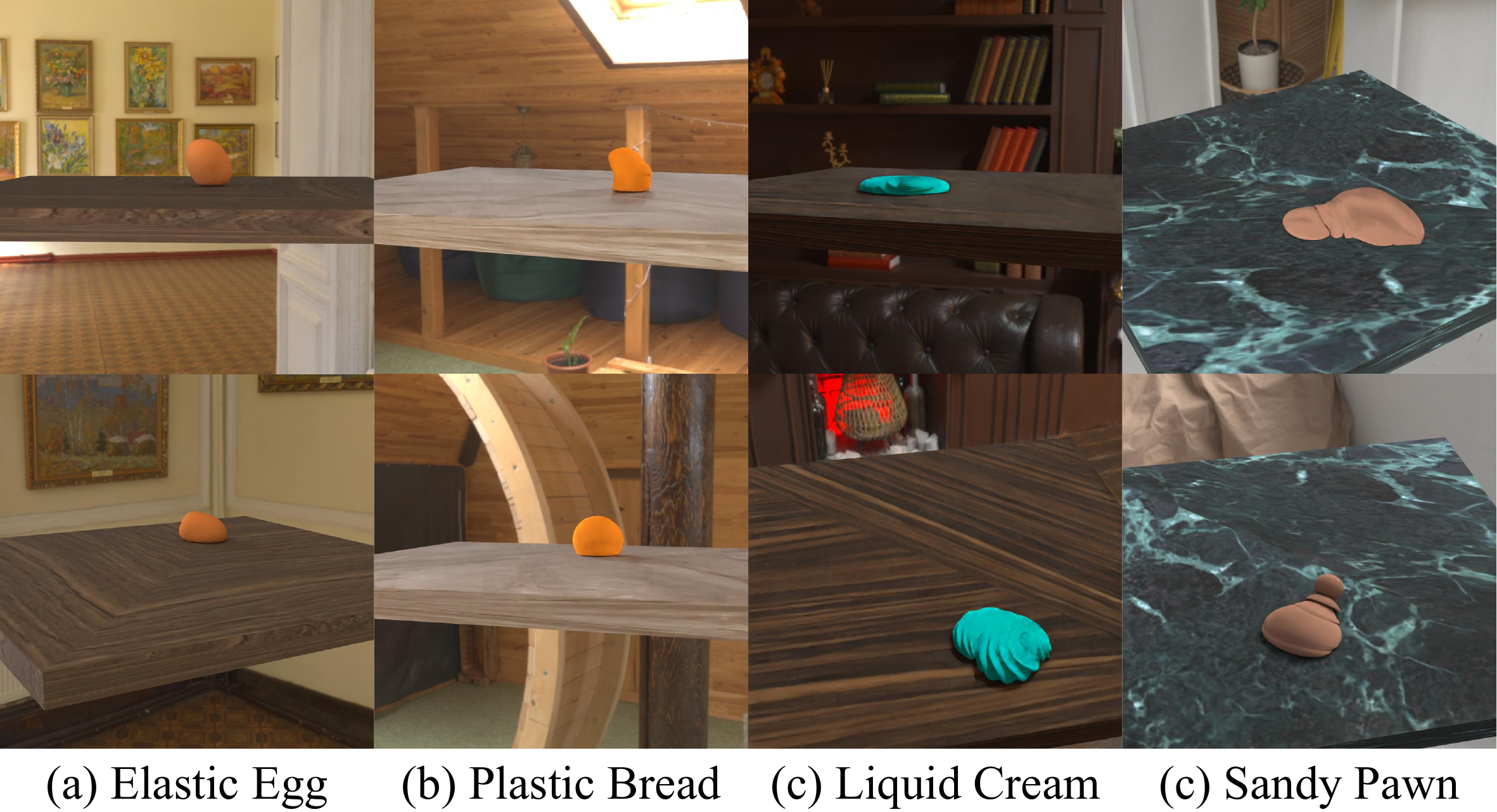}
    \caption{\textbf{The multi-sequence dataset synthesized with Genesis~\cite{Genesis}.} It includes 4 objects in 4 multi-view sequences simulated under randomized initial locations, poses, and velocities.}
\label{fig:multi_sequence}
\end{figure}

\section{More Quantitative Analysis}
\label{sec:supp:quantitative}
\begin{table}[t]
\centering
\footnotesize
\setlength{\tabcolsep}{0.3em}
\adjustbox{width=\linewidth}{
\begin{tabular}{llc|cccca}
\toprule
    & Method & \# Train & Elastic Egg & Plastic Bread & Liquid Cream & Sandy Pawn & Mean \\
\midrule
    & & 1 & 2.10 & 0.66 & 0.81 & 0.20 & 0.94 \\
    & GIC~\cite{cai2024gic} & 2 & 2.33 & 1.40 & 0.12 & 0.20 & 1.01 \\
    & & 3 & 2.25 & 2.98 & \textbf{0.11} & 0.20 & 1.39 \\
    \rowcolor{Gray}\cellcolor{white} & & 1 & 11.36 & 89.42 & 15.65 & 0.55 & 29.24 \\
    \rowcolor{Gray}\cellcolor{white} & Ours & 2 & 4.68 & 5.42 & 1.81 & 0.15 & 3.02 \\
    \rowcolor{Gray}\cellcolor{white} \multirow{-6}{*}{\rotatebox{90}{CD~$\downarrow$}} & & 3 & \textbf{1.78} & \textbf{0.28} & 0.12 & \textbf{0.11} & \textbf{0.57} \\
\midrule
    & & 1 & 35.24 & 32.29 & 38.38 & 32.34 & 34.56 \\
    & GIC~\cite{cai2024gic} & 2 & 34.87 & 30.94 & 41.31 & 32.37 & 34.87 \\
    & & 3 & \textbf{34.99} & 30.45 & 41.52 & 32.38 & 34.83 \\
    \rowcolor{Gray}\cellcolor{white} & & 1 & 28.54 & 27.58 & 36.64 & 31.69 & 31.11 \\
    \rowcolor{Gray}\cellcolor{white} & Ours & 2 & 30.25 & 28.67 & 38.66 & 33.65 & 32.81 \\
    \rowcolor{Gray}\cellcolor{white} \multirow{-6}{*}{\rotatebox{90}{PSNR~$\uparrow$}} & & 3 & 30.48 & \textbf{33.73} & \textbf{41.62} & \textbf{35.22} & \textbf{35.26} \\
\midrule
    & & 1 & \textbf{0.992} & 0.992 & 0.992 & 0.991 & \textbf{0.992} \\
    & GIC~\cite{cai2024gic} & 2 & 0.991 & 0.990 & 0.994 & 0.991 & \textbf{0.992} \\
    & & 3 & \textbf{0.992} & 0.988 & \textbf{0.995} & 0.991 & 0.991 \\
    \rowcolor{Gray}\cellcolor{white} & & 1 & 0.980 & 0.982 & 0.987 & 0.992 & 0.985 \\
    \rowcolor{Gray}\cellcolor{white} & Ours & 2 & 0.985 & 0.985 & 0.992 & 0.993 & 0.989 \\
    \rowcolor{Gray}\cellcolor{white} \multirow{-6}{*}{\rotatebox{90}{SSIM~$\uparrow$}} & & 3 & 0.987 & \textbf{0.993} & \textbf{0.995} & \textbf{0.994} & \textbf{0.992} \\
\bottomrule
\end{tabular}}
\caption{\textbf{Inter-sequence generalization on multi-sequence dataset.} \model improves performance with more training data.}
\label{tab:multi_sequence}
\vspace{-0.6em}
\end{table}

In~\cref{tab:multi_sequence}, we examine the advantages of using a data-driven constitutive model instead of predefined ones.
Specifically, we generate multi-view videos for 4 objects of different materials with Genesis~\cite{Genesis}, each including 4 multi-view sequences simulated under randomized initial conditions varying in location, pose, and velocity.
The multi-view cameras are set to be the same as those in the PAC-NeRF dataset.
Some samples of this synthetic dataset are shown in~\cref{fig:multi_sequence}.
We progressively add the number of training sequences for system identification and test the estimated parameters on an unseen sequence.
The results show that \model improves the estimation accuracy with more observations of a certain object, highlighting its data-driven advantages.
In contrast, GIC~\cite{cai2024gic} exhibits relatively similar performance regardless of data quantity, likely due to its reliance on fixed constitutive priors.

\section{More Qualitative Comparison}
\label{sec:supp:qualitative}
\begin{figure*}[h]
    \centering
    \includegraphics[width=0.75\linewidth]{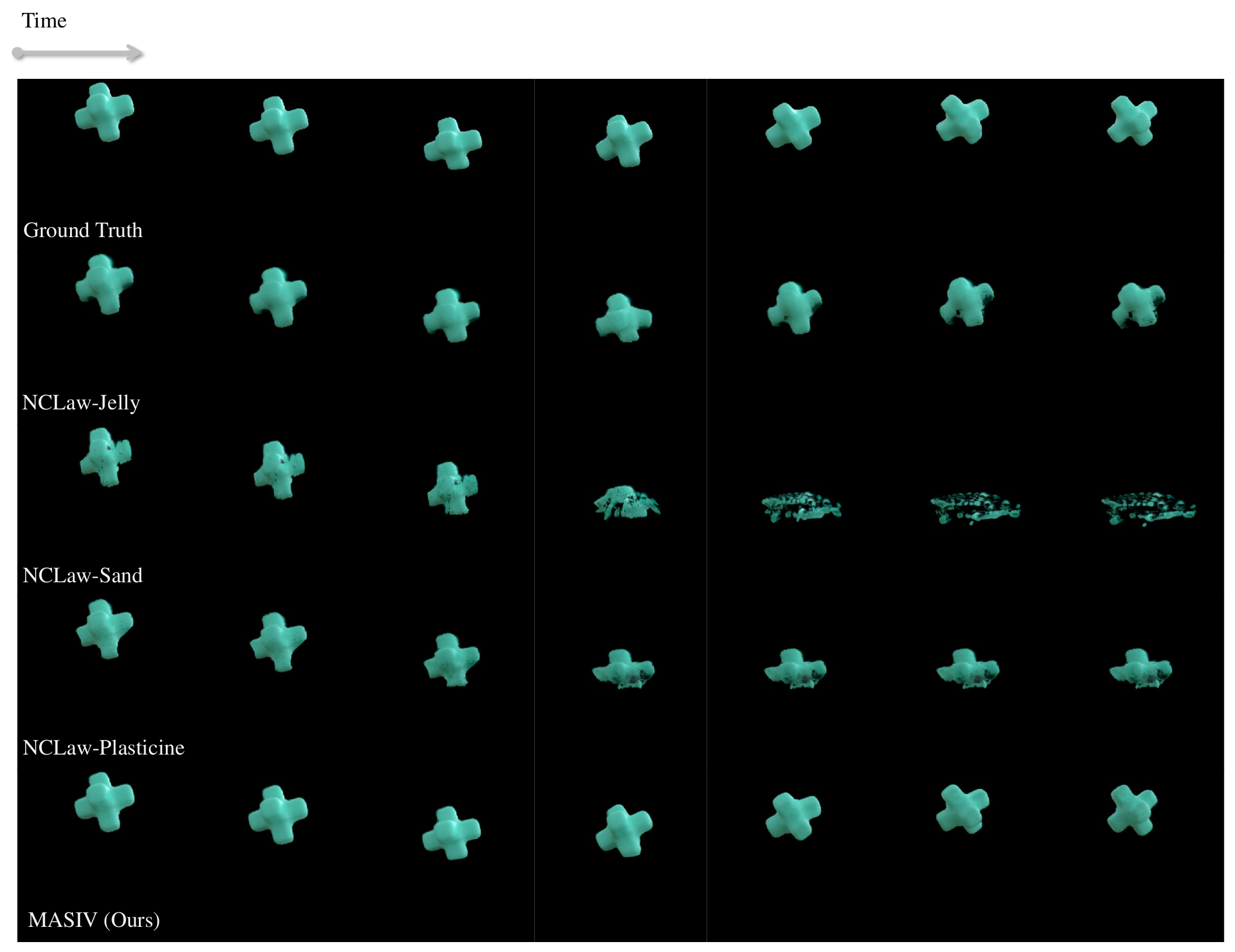}
    \caption{\textbf{Qualitative comparisons on the cross-shaped Elastic subset of the PAC-NeRF Dataset.}}
    \label{fig:qualitative_pretrained_finetuned_elastic}
\end{figure*}

\begin{figure*}[h]
    \centering
    \includegraphics[width=0.75\linewidth]{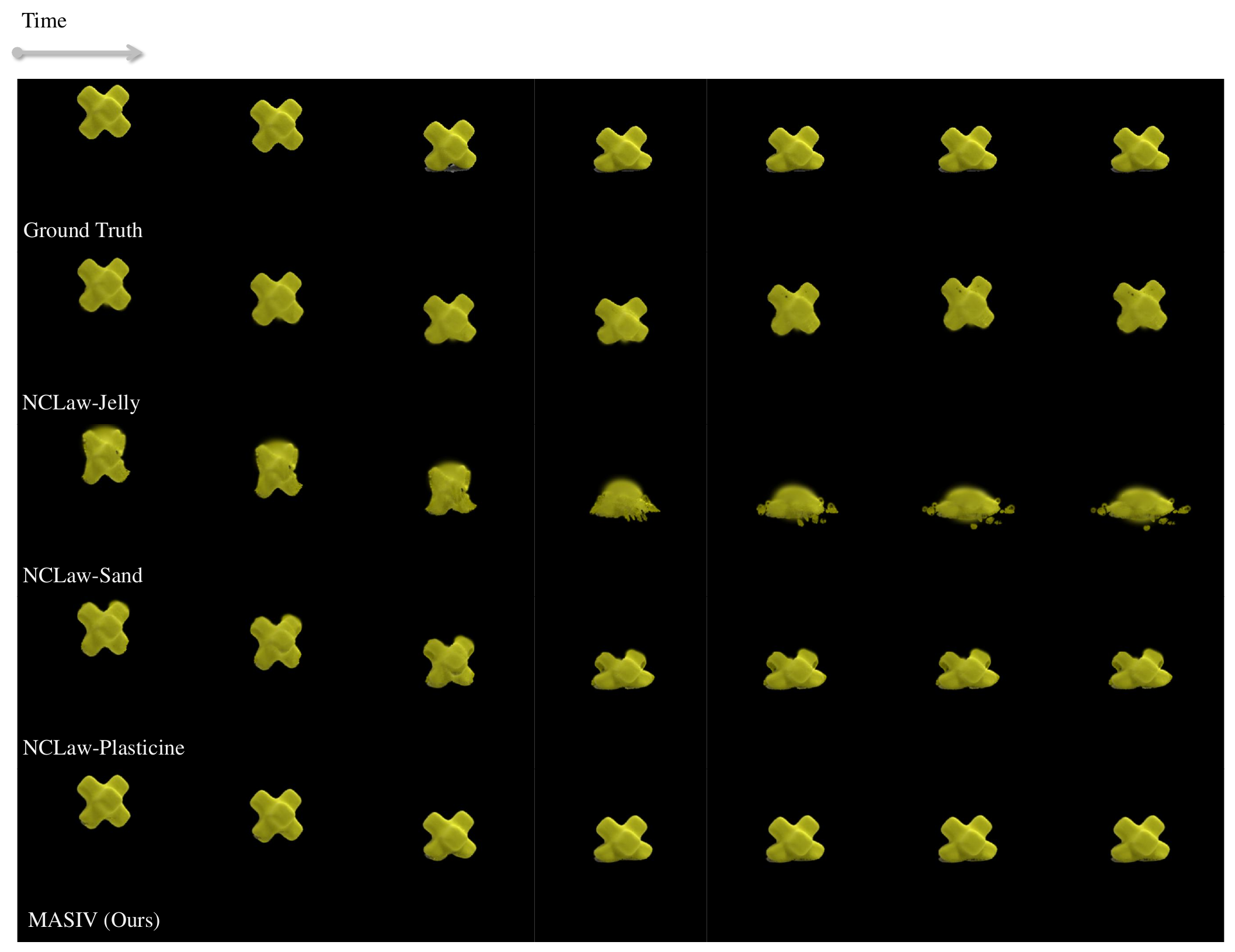}
    \caption{\textbf{Qualitative comparisons on the cross-shaped Plasticine subset of the PAC-NeRF Dataset.}}
    \label{fig:qualitative_pretrained_finetuned_plasticine}
\end{figure*}

\begin{figure*}[h]
    \centering
    \includegraphics[width=0.75\linewidth]{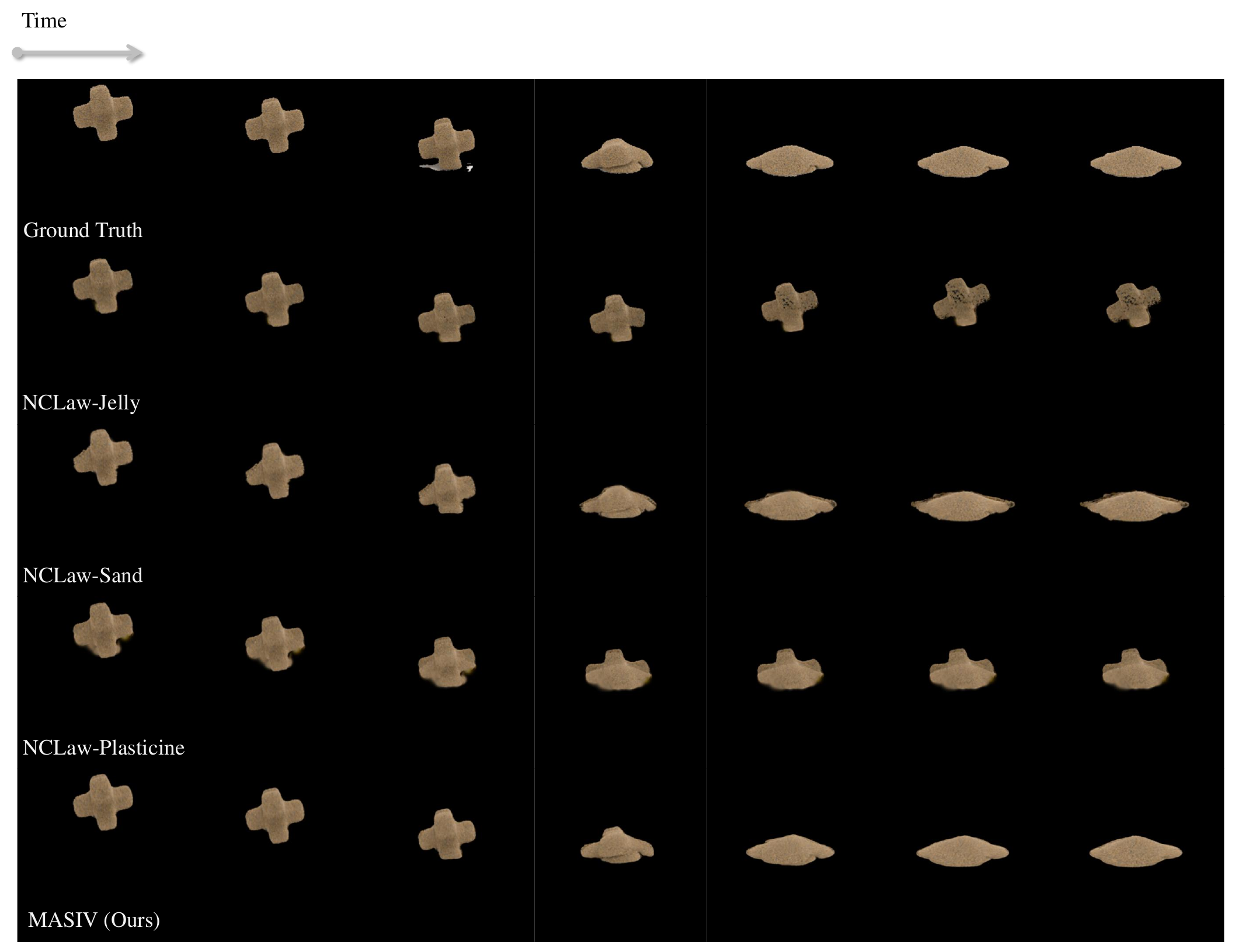}
    \caption{\textbf{Qualitative comparisons on the cross-shaped Sand subset of the PAC-NeRF Dataset.}}
    \label{fig:qualitative_pretrained_finetuned_sand}
\end{figure*}

\begin{figure*}[h]
    \centering
    \includegraphics[width=0.75\linewidth]{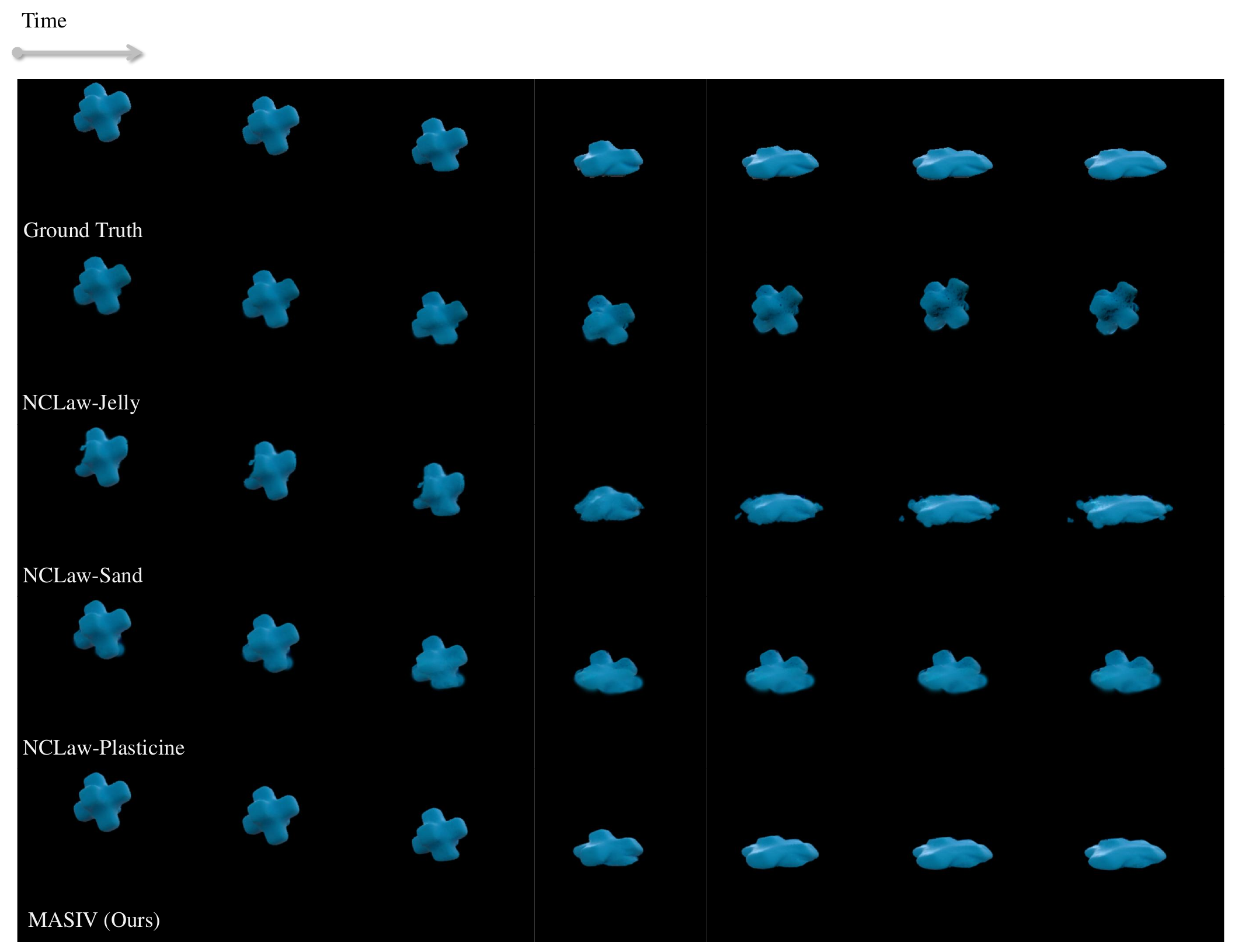}
    \caption{\textbf{Qualitative comparisons on the cross-shaped Newtonian subset of the PAC-NeRF Dataset.}}
    \label{fig:qualitative_pretrained_finetuned_newtonian}
\end{figure*}

\begin{figure*}[h]
    \centering
    \includegraphics[width=0.75\linewidth]{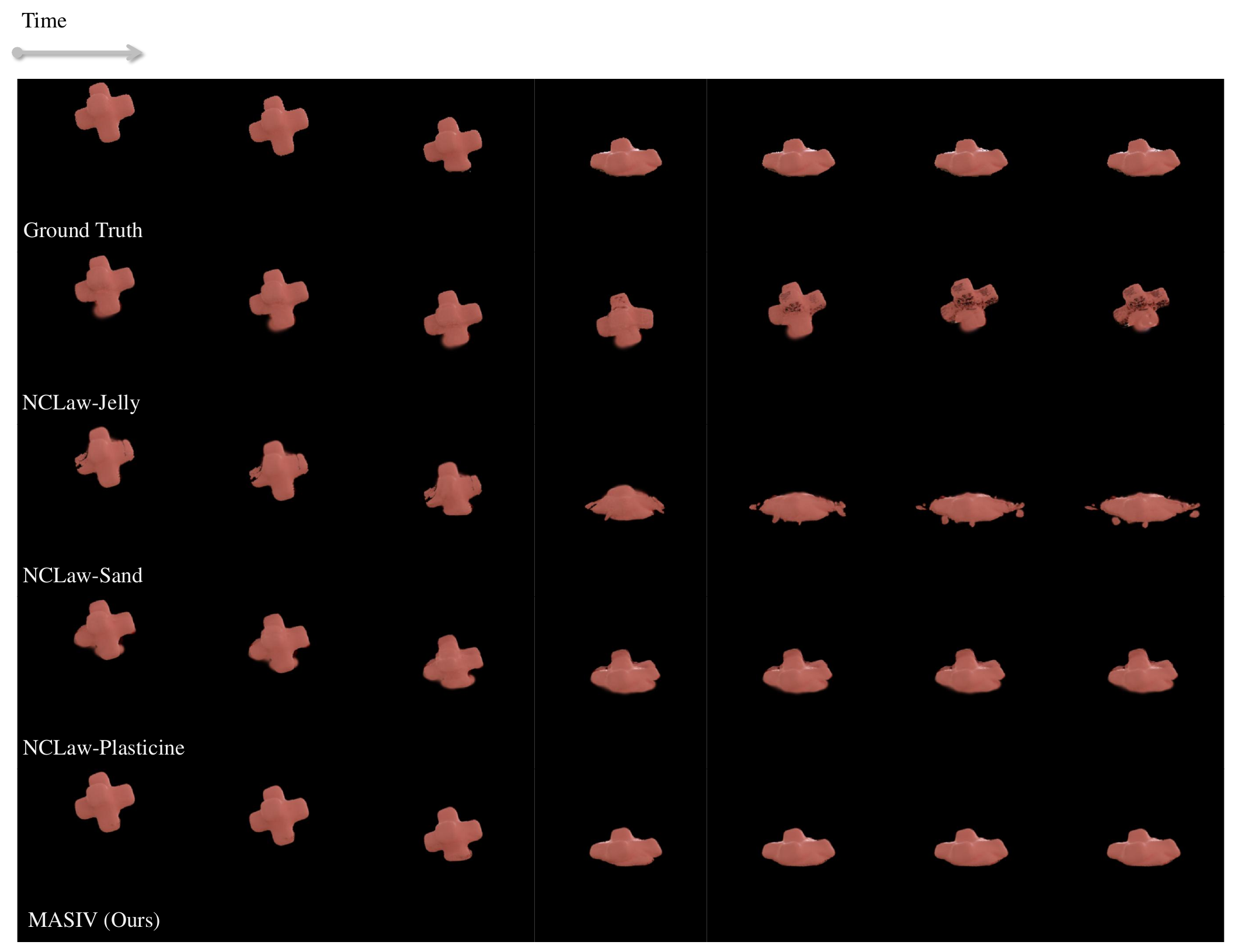}
    \caption{\textbf{Qualitative comparisons on the cross-shaped Non-Newtonian subset of the PAC-NeRF Dataset.}}
    \label{fig:qualitative_pretrained_finetuned_non_newtonian}
\end{figure*}

We conduct extra qualitative comparison on multi-material cross-shaped subsets of the PAC-NeRF dataset, showing the results from NCLaw~\cite{ma2023learning} with full-state data pre-training and our \model in~\cref{fig:qualitative_pretrained_finetuned_elastic,fig:qualitative_pretrained_finetuned_plasticine,fig:qualitative_pretrained_finetuned_sand,fig:qualitative_pretrained_finetuned_newtonian,fig:qualitative_pretrained_finetuned_non_newtonian}.
Each figure compares the results of multiple approaches, including NCLaw pre-trained on Jelly, Plasticine, or Sand, alongside our \model fine-tuned over NCLaw-Jelly, against the ground truth over time.
From the qualitative comparisons, it is evident that \model consistently demonstrates improved fidelity to the ground truth across all material types.
For elastic objects (\cref{fig:qualitative_pretrained_finetuned_elastic}), \model captures finer deformations with greater accuracy.
For sandy objects (\cref{fig:qualitative_pretrained_finetuned_sand}), \model better preserves granular motion.
In plasticine (\cref{fig:qualitative_pretrained_finetuned_plasticine}), Newtonian (\cref{fig:qualitative_pretrained_finetuned_newtonian}), and non-Newtonian (\cref{fig:qualitative_pretrained_finetuned_non_newtonian}) objects, \model maintains structure and flow consistency, demonstrating adaptability to diverse material properties.
These results emphasize the ability of \model to ground intrinsic dynamics with visual clues, without the need for an initialization sufficiently close to the material of interest.

